\documentclass{article}

\usepackage{microtype}
\usepackage{graphicx}
\usepackage{subcaption}
\usepackage{booktabs} 

\usepackage{hyperref}

\usepackage[accepted]{icml2026}
\usepackage{amsmath}
\usepackage{amssymb}
\usepackage{mathtools}
\usepackage{amsthm}
\usepackage[capitalize,noabbrev]{cleveref}
\usepackage{hyperref}
\usepackage{url}
\usepackage{amsfonts}
\usepackage{amsthm}
\usepackage{xcolor}         
\usepackage{dsfont}
\usepackage{wrapfig}
\usepackage{threeparttable}
\usepackage{algorithm} 
\usepackage{algorithmic} 
\usepackage{multirow}
\usepackage{mdframed}
\usepackage{enumitem}
\usepackage{soul}

\theoremstyle{plain}
\newtheorem{theorem}{Theorem}
\newtheorem{proposition}{Proposition}
\newtheorem{lemma}{Lemma}
\newtheorem{corollary}{Corollary}
\newtheorem{property}{Property}
\theoremstyle{definition}
\newtheorem{definition}{Definition}

\theoremstyle{remark}
\newtheorem{remark}{Remark}
\usepackage[textsize=tiny]{todonotes}

\newcommand{\apxAauto}[3]{%
  \noindent\makebox[\linewidth][r]{%
    \begin{tabular*}{0.96\linewidth}{@{}p{0.94\linewidth}@{\extracolsep{\fill}}r@{}}%
      \makebox[2.0em][l]{\textbf{#1}}%
      \hyperref[#3]{\textbf{#2}\dotfill}%
      & \hyperref[#3]{\pageref*{#3}}\\%
    \end{tabular*}%
  }%
  \vspace{0.3em}%
}

\newcommand{\apxAoneauto}[3]{%
  \noindent\makebox[\linewidth][r]{%
    \begin{tabular*}{0.96\linewidth}{@{}p{0.94\linewidth}@{\extracolsep{\fill}}r@{}}%
      \hspace*{3em}%
      \makebox[4.0em][l]{#1}%
      \hyperref[#3]{#2\dotfill}%
      & \hyperref[#3]{\pageref*{#3}}\\%
    \end{tabular*}%
  }%
  \vspace{0.2em}%
}

\icmltitlerunning{Sample-Level Constraint Relaxation for Offline-to-Online RL}

\begin{document}

\twocolumn[
  \icmltitle{From Static Constraints to Dynamic Adaptation: Sample-Level Constraint Relaxation for Offline-to-Online Reinforcement Learning}
  \begin{icmlauthorlist}
    \icmlauthor{Lipeng Zu}{1}
    \icmlauthor{Yu Qian}{1}
    \icmlauthor{Shayok Chakraborty}{1}
    \icmlauthor{Xiaonan Zhang}{1}
  \end{icmlauthorlist}
  \icmlaffiliation{1}{Department of Computer Science, Florida State University, Tallahassee, FL, USA}
  \icmlcorrespondingauthor{Xiaonan Zhang}{xzhang14@fsu.edu}
  \icmlkeywords{Machine Learning, ICML}
  \vskip 0.3in
]

\printAffiliationsAndNotice{}  

\begin{abstract}
Offline-to-online reinforcement learning (O2O RL) faces a central challenge between retaining offline conservatism and adapting to online feedback under distribution shift.  This  challenge arises because data behavior  evolves  during fine-tuning,  rendering  data origin a misleading basis for constraint handling and thereby leading to objective–data mismatch. We therefore propose Dynamic Alignment for RElaxation (DARE), a distribution-aware framework for sample-level constraint relaxation based on the behavioral consistency with a behavior model. To our knowledge, DARE is the first to condition constraint relaxation on behavioral consistency via a posterior-induced exchange mechanism, moving beyond a binary offline/online data distinction. Importantly, DARE requires only per-sample behavioral alignment, enabling instantiation on top of many offline algorithms with flexible choices of behavior models and fine-tuning objectives. We provide a theoretical analysis showing that behavior-based sample exchange consistently improves the distinction between offline-like and online-like subsets. Experiments on D4RL demonstrate that DARE consistently improves fine-tuning stability and achieves superior final performance over strong offline-to-online baselines.
(The code is publicly available at \url{https://github.com/lpzu/DARE}.)
\end{abstract}

\section{Introduction}
Offline-to-online reinforcement learning (O2O RL) aims to combine the efficiency of offline pretraining with the adaptability of online fine-tuning~\citep{xie2021policy, zhang2024policy}. By initializing policies from offline dataset~\citep{fujimoto2019off, kumar2020conservative, uehara2024bridging} and refining them through limited online interactions~\citep{mnih2015human, song2023reaching, alonso2024diffusion}, O2O RL provides a practical pathway toward scalable and safe RL in complex tasks~\citep{lee2022offline, he2025robust}. However, offline pretraining typically relies on conservative constraints to ensure stability in the absence of online feedback~\citep{ball2023efficient}. Once the agent enters online fine-tuning, a distribution shift naturally arises because the evolving policy induces behaviors that differ from those represented in the offline dataset~\citep{an2021uncertainty, yang2022rorl}. As a result, conservative constraints can conflict with the need for exploration and adaptation~\citep{yu2023actor}. This inherent  discrepancy lies at the central challenge of the O2O setting.

To cope with this fundamental challenge, most existing O2O approaches regulate policy updates or value estimation to preserve stability~\citep{wu2022supported, zhang2023policy}. In practice, this regulation is achieved by explicitly removing conservative constraints inherited from offline training or incorporating them through soft transition mechanisms during online fine-tuning~\citep{wang2023train, xu2026uni}. Despite these design choices, these methods typically apply uniform training objectives across samples from different data origins~\citep{luo2024optimistic, mao2024doubly}. This uniformity thereby obfuscates the distinction between samples that are meant to anchor conservative calibration and those that are meant to support online adaptation. 

A seemingly natural response is to treat data differently based on its origin, followed by applying different training objectives accordingly~\citep{zheng2023adaptive, shin2025online}. This strategy uses data origin as the basis for assigning samples to different constraint rules. However, the distribution shift that matters in O2O RL is characterized by behavior relative to the state–action support of the offline dataset, which cannot be inferred from collection origin alone. As fine-tuning proceeds, both the offline dataset and newly collected online data can contain a mixture of offline-like and online-like behaviors. Consequently, separating data by origin can still lead to the same objective–data misalignment as uniform training. This observation motivates a sample-level approach to constraint relaxation that relies on behavioral consistency rather than data origin.

Building on the above analysis, we highlight three design requirements for constraint relaxation decision: $(i)$ the decision criterion is behavior-driven rather than origin-driven; $(ii)$ the decision granularity operates at the sample level to reflect heterogeneous roles within mixed data; and $(iii)$ the decision rule is explicit and consistent within a well-defined formulation. To satisfy these requirements, we propose Dynamic Alignment for RElaxation (DARE), a distribution-aware framework for sample-level constraint handling. DARE evaluates sample-level behavioral alignment through a Bayes-optimal posterior. The posterior induced by the quadratic structure determines which samples can be exchanged between conservative and adaptive regularization, yielding a unified exchange rule for constraint-relaxation decisions.

We theoretically derive the posterior-induced exchange mechanism and demonstrate that it enhances the separability between offline-like and online-like behaviors. Empirically, we instantiate DARE within Cal-QL~\citep{nakamoto2023cal} and IQL~\citep{kostrikov2022offline}, achieving consistent performance gains over the corresponding baselines on standard benchmarks. Additional experiments and ablations on different instantiations further validate the effectiveness of the DARE framework.

\section{Related Work}
\noindent\textbf{Constraint Handling in O2O RL.}
Offline-to-online RL aims to combine reliable value estimation from offline data with adaptive policy learning during online fine-tuning under distributional shift~\citep{he2025robust, li2025reinforcement}. Some approaches, such as FamO2O~\citep{wang2023train} and Uni-RL~\citep{xu2026uni}, adopt a uniform objective formulation throughout training to avoid abrupt changes in learning signals. Other methods focus on stabilizing training dynamics: Cal-QL~\citep{nakamoto2023cal} calibrates Q-values during offline training, whereas several online approaches~\citep{zhang2024perspective,feng2024suf,ball2023efficient} further stabilize value learning. Another line of work addresses the instability of ensemble Q-functions after offline-to-online transfer~\citep{guo2026tackling, hu2024bayesian}, where critic disagreement or biased value aggregation can be amplified by newly collected online data. However, these methods primarily keep stabilization via uniform training objective, without explicitly distinguishing offline and online data. In contrast, other approaches explicitly separate optimization objectives by data origin~\citep{zheng2023adaptive}. For example, OPT~\citep{shin2025online} retrains the critic with online-style objectives applied based on data origin and accelerates adaptation by increasing the update-to-data (UTD) ratio. Nevertheless, such separations are usually defined at the level of data origin, yielding coarse-grained assignments that overlook sample-level behavioral heterogeneity.

\noindent\textbf{Behavior Modeling and Distribution Alignment.~}
Behavior modeling has been widely explored as a approach to providing a reference for aligning learning dynamics with offline data. To model behavioral imitation in a generative manner, diffusion-based behavior modeling has been explored as a generative prior over actions or trajectories~\citep{janner2022planning, chen2023offline, lu2023synthetic}. Within this line of work, CEP~\citep{lu2023contrastive} approximates energy-based guidance through contrastive learning, primarily focusing on offline behavior generation, while EDIS~\citep{liu2024energy} extends energy-guided diffusion to the O2O setting by steering action sampling during fine-tuning. In contrast to approaches that explicitly model behavior~\citep{rigter2024world}, imitation-regularized methods such as AWAC \citep{nair2020awac} applies advantage-weighted imitation to keep policy updates aligned with offline behaviors, while PEX \citep{zhang2023policy} preserves offline-consistent behaviors during online adaptation through an offline-trained policy. Separately, Off2On \citep{lee2022offline} interpolates between conservative offline pretraining and exploratory online fine-tuning via replay reweighting. Across these methods, behavioral components are predominantly used for regularization or generation, without being explicitly designed to guide sample-level constraint handling.

Unlike prior work that assigns objectives by training phase or data origin, we leverage behavioral consistency relative to a learned behavior model to make sample-level constraint-handling decisions.

\section{Preliminaries}
\subsection{Reinforcement Learning}
RL is a framework in which an agent learns to maximize cumulative rewards by interacting with an environment~\citep{mnih2013playing,van2016deep}. The problem is  modeled as a Markov Decision Process (MDP), defined by a tuple $(\mathcal{S}, \mathcal{A}, P, R, \gamma)$, where $\mathcal{S}$ is the state space, $\mathcal{A}$ is the action space, $P$ is the transition probability, $R$ is the reward function, and $\gamma \in [0, 1)$ is the discount factor.

\subsection{Offline-to-online RL}
Offline-to-online (O2O) RL commonly follows a two-phase training paradigm~\citep{hu2024bayesian,zheng2025value}: (i) \emph{offline pretraining}, where a policy is learned from a fixed dataset $\mathcal{D}_{\text{off}}$ without interacting with the environment; and (ii) \emph{online fine-tuning}, where the pre-trained policy interacts with the environment, and the newly collected transitions are stored in an online replay buffer $\mathcal{D}_{\text{on}}$. During online fine-tuning, each update typically uses a mini-batch $b$ mixing samples from both buffers, which is
\(
b = b_{\text{off}} \cup b_{\text{on}}.
\)

\subsection{Constraints in O2O RL}
Conservative constraints are widely used in offline RL to stabilize value estimation under distribution shift. During the offline-to-online transition, constraint handling often involves a trade-off between enabling online adaptation and maintaining stable policy improvement. In many algorithms, this is encoded through the learning objective. We outline the two representative classes below
\begin{itemize}[leftmargin=0.15in, itemsep=1.pt]
  \item \textbf{Explicit conservatism.}
  Conservative Q-Learning (CQL) explicitly regularizes the Q-function to discourage overestimation on actions that are weakly supported by the data distribution~\citep{kumar2020conservative}. In general, the explicit conservatism can be implemented by augmenting the Bellman-error objective with a regularizer \(\mathcal{R}\)
  \[
  \resizebox{0.98\linewidth}{!}{$
    \mathcal{L}_{\text{CQL}}(Q) = \mathbb{E}_{(s,a,s') \sim \mathcal{D}} \left[ \left( Q(s,a) - \hat{\mathcal{B}}^{\pi} \hat{Q}_{\text{target}}(s,a) \right)^2 \right] + \alpha \mathcal{R},
  $}
  \]
  where $\alpha$ controls the strength of conservatism.

  \item \textbf{Implicit conservatism.}
  Implicit Q-Learning (IQL) induces conservatism implicitly through expectile-based value learning, discouraging the exploitation of unseen actions~\citep{kostrikov2022offline}. Briefly, IQL first learns an implicit value function through expectile regression 
  \[
    \mathcal{L}_{IQL}(V) = \mathbb{E}_{(s,a) \sim \mathcal{D}} \left[ L_{\tau}^{2}\left(\hat{Q}_{\text{target}}(s, a) - V(s)\right) \right],
  \]
  where $L_{\tau}^{2}(u) = |\tau - \mathds{1}{(u < 0)}| u^2$ is the expectile regression loss. The learned value function is then used to update the Q-function and corresponding actor function.
\end{itemize}
CQL and IQL can both be viewed as mechanisms for enforcing conservatism within value-based policy optimization. In this paper, we use them to demonstrate and validate our approach. More broadly, the proposed approach is generic and can be integrated into a wide range of O2O RL.

\section{Motivation}
Most existing O2O RL methods fine-tune on mixed offline and online data using a single optimization objective. As the behavior distribution changes during fine-tuning, a unified objective  struggles to balance training stability with online adaptation. A straightforward approach is to employ different optimization objectives for offline and online data respectively, thereby decoupling conservative calibration from online improvement. However, this approach still relies on the assumption that offline and online data are behaviorally separable. To examine the validity of this assumption under realistic fine-tuning, we conduct a diagnostic experiment.

\begin{figure}[ht]
  \centering
  \begin{subfigure}{0.23\textwidth}
    \includegraphics[width=\linewidth]{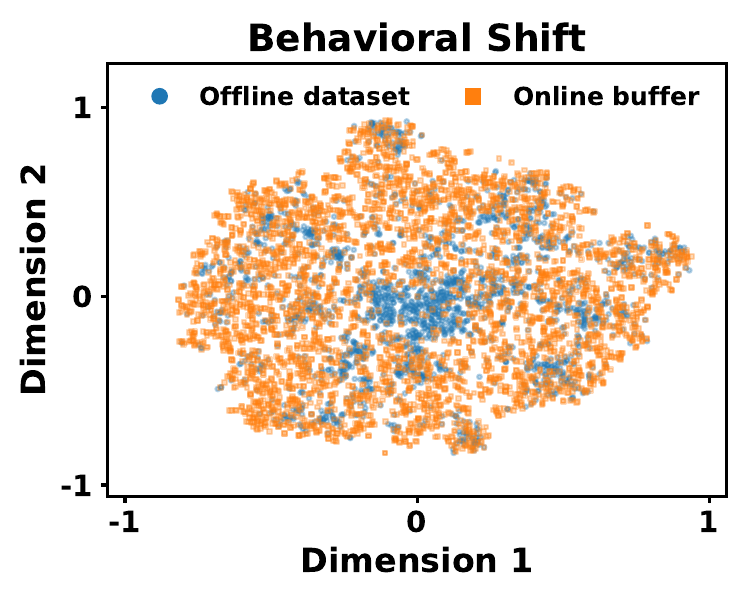}
    \caption{t-SNE visualization.}
    \label{fig:motivation_a}
  \end{subfigure}
  \begin{subfigure}{0.23\textwidth} 
  \centering
    \includegraphics[width=\linewidth]{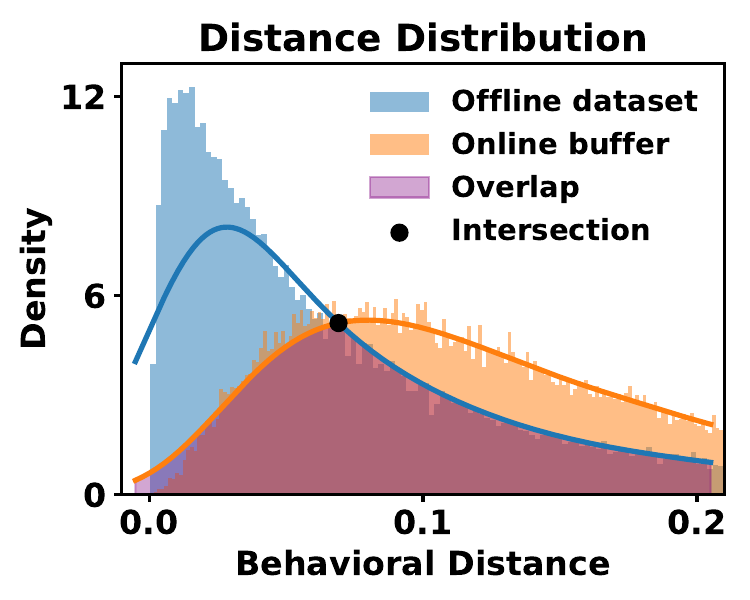}
    \caption{Distance distribution.}
    \label{fig:motivation_b}
\end{subfigure}
\caption{Characteristics of behavior shift in fine-tuning.}
\end{figure}

We study the HalfCheetah environment~\citep{towers2026gymnasium} under the O2O setting by collecting a $100$k-size online replay buffer. We then compare this buffer to the offline dataset using an offline behavior imitation model, which serves as a behavioral reference for cross-dataset comparison. As shown in Fig.~\ref{fig:motivation_a}, offline samples concentrate in a central region of the learned representation space, whereas online samples are more dispersed. Notably, there is a nontrivial overlap between offline and online samples, demonstrating that offline and online data are not strictly separable in the behavior space.

To quantify the overlap between offline and online data, we compute, for each sample, its behavioral distance to the offline imitation model. The resulting distributions in Fig.~\ref{fig:motivation_b} exhibit a clear intersection between offline and online samples. Rather than forming disjoint groups, the two datasets occupy overlapping regions along a continuous behavioral axis. As a result, assigning optimization objectives solely based on data origin inevitably misaligns with the behavioral roles of individual samples. In particular, samples near the overlap region cannot be reliably assigned a fixed constraint role, since they may support either conservative calibration or online adaptation depending on their behavioral alignment. This observation motivates a sample-level mechanism that dynamically reassigns constraint roles, rather than enforcing static, origin-based distinctions.

\section{DARE: Dynamic Alignment for RElaxation}
We propose DARE, an O2O RL framework built around a posterior-induced exchange mechanism for sample-level constraint handling. The exchange mechanism leverages behavioral alignment to infer whether each sample is offline-like or online-like, and reassigns constraint roles accordingly. This preserves conservative constraints for offline-like samples while selectively relaxing them for samples exhibiting online-like behavior. We further show that the exchange mechanism admits a Bayes-optimal posterior interpretation and provably improves offline--online behavioral separability. All formal proofs are deferred to \textbf{Appendix~\ref{appx:proofs}}.

\subsection{Bayes--Optimal Posterior}
We consider samples drawn from an offline dataset and an online replay buffer, indexed by a binary class label $C\in\{0,1\}$. Let $(s,a)\in\mathcal{S}\times\mathcal{A}$ denote a state--action pair, and let $\pi_b$ be a reference behavior model. We define a scalar behavioral deviation statistic
\begin{equation}
\label{eq:distance}
d(s,a) := \operatorname{dist}\bigl(a, a_b(s)\bigr),
\end{equation}
where $\operatorname{dist}(\cdot,\cdot)$ is a deviation measure that quantifies behavioral discrepancy and $a_b(s)$ denotes a reference action induced by $\pi_b$. We instantiate $\operatorname{dist}(\cdot,\cdot)$ as the $\ell_2$ distance.

Let \(X := A \mid S\) denote the state-conditioned action variable from a behavioral perspective. We model the source discrimination being governed by behavioral deviation relative to the reference model, formalized through the following likelihood--ratio structure.

\begin{theorem} 
\label{thm:disc-suff} 
Suppose the class-conditional densities admit a likelihood--ratio representation through \(d\); that is, there exists a measurable function \(r:\mathbb{R}\to(0,\infty)\) such that
\[
\frac{p(X=x\mid C=0)}{p(X=x\mid C=1)} = r\!\bigl(d(s,a)\bigr).
\]

It follows that \(d\) is sufficient for Bayes--optimal posterior inference between the two classes, i.e. there exists a measurable function \(g:\mathbb{R}\to[0,1]\) such that
\[
\mathbb{P}(C=0 \mid X=x) = g\!\bigl(d(s,a)\bigr).
\]
\end{theorem}

Thm.~\ref{thm:disc-suff} justifies reducing the source-discrimination problem from the high-dimensional behavioral space to the scalar statistic \(D=d(s,a)\). Specifically, \(r(\cdot)\) appears on the assumption side by relating the likelihood ratio to the scalar \(d\), whereas \(g(\cdot)\) appears on the conclusion side by representing the posterior as a function of the same statistic. Since \(D\) is a measurable statistic of the variable \((S,A)\), it induces a sub-\(\sigma\)-algebra of the full state--action space
\[
\sigma(D)=\sigma_b(S,A)\subseteq\sigma(S,A),
\]
where \(\sigma_b(S,A)\) denotes the behavior-relevant sub-\(\sigma\)-algebra retained by the scalar statistic \(D\). Thus, function $g$ in Thm.~\ref{thm:disc-suff} is given by
\begin{equation}
    g(d(s,a))=\mathbb{P}(C=0\mid D=d).
    \label{eq:g}
\end{equation}

To obtain a tractable posterior form, we further model the class--conditional distribution of $D$ as Gaussian
\[
D\mid C=c \sim \mathcal{N}(\mu_c,\sigma_c^2),\qquad c\in\{0,1\}.
\]
Let $\rho_c:=\mathbb{P}(C=c)$ denote the class priors, with $\rho_0,\rho_1>0$ and $\rho_0+\rho_1=1$. By Bayes' rule, the posterior probability in Eq.~\ref{eq:g} admits the closed-form expression
\begin{equation}
\label{eq:posterior}
\resizebox{0.91\linewidth}{!}{$
\begin{aligned}
\mathbb{P}(C=0 \mid D=d)
&=\frac{\rho_0\,p(d\mid C=0)}
{\rho_0\,p(d\mid C=0)+\rho_1\,p(d\mid C=1)}\\
&=\frac{1}{1+\frac{\rho_1\,\sigma_0}{\rho_0\,\sigma_1}
\exp\!\left(-\frac{(d-\mu_1)^2}{2\sigma_1^2}
+\frac{(d-\mu_0)^2}{2\sigma_0^2}\right)}.
\end{aligned}
$}
\end{equation}
This posterior form serves as the basis for deriving symmetry-induced thresholds and the exchange rules designed in the subsequent sections.

\begin{remark}
\label{thm:remark}
Although the distance statistic $d$ is nonnegative, modeling its class-conditional distribution as Gaussian can be justified in a local approximation sense. For a broad class of positive-support distributions (e.g., Gamma), the log-density admits a second-order Taylor expansion around regions of high probability mass, yielding a quadratic form in $d$. This local quadratic approximation implies that, around a local maximizer of the log-density, the class-conditional log-density is well approximated by that of a Gaussian model. A more detailed analysis of the Gaussian assumption is provided in \textbf{Appendix~\ref{appx:gauss}}.
\end{remark}

\subsection{Symmetry-Induced Thresholds}
We model the behavioral statistic $D$ using a two-class Gaussian mixture, which allows us to compute the posterior probability via the Bayes' rule. Although the posterior takes a logistic form, its qualitative behavior is entirely governed by the quadratic term in the likelihood ratio. Specifically, the posterior in Eq.~\ref{eq:posterior} can be re-written as
\begin{equation}
\label{eq:f}
\mathbb{P}(C=0 \mid D=d)
=
\frac{1}{1+\alpha \exp\!\big(\ell(d)\big)},
\end{equation}
where $\alpha=\frac{\rho_1\sigma_0}{\rho_0\sigma_1}>0$ and
\[
\ell(d)
=
-\frac{(d-\mu_1)^2}{2\sigma_1^2}
+\frac{(d-\mu_0)^2}{2\sigma_0^2}
=
\ell_1 d^2 + \ell_2 d + \ell_3.
\]
Since the logistic function is strictly monotonic, the monotonicity and extremal behavior of the posterior in Eq.~\ref{eq:f} are fully determined by the quadratic function $\ell(d)$. When $\sigma_0 \neq \sigma_1$, $\ell(d)$ admits a unique stationary point
\begin{equation}
\label{eq:d}
d^\star = -\frac{\ell_2}{2\ell_1}
=
\frac{\mu_1/\sigma_1^2 - \mu_0/\sigma_0^2}
{1/\sigma_1^2 - 1/\sigma_0^2}.
\end{equation}

\begin{theorem}
\label{thm:dstar-outside}
The point $d^\star$ in Eq.~\ref{eq:d} satisfies $d^\star\notin(\mu_0,\mu_1)$.
\end{theorem}

As a consequence, the posterior $\mathbb{P}(C=0 \mid D=d)$ remains strictly monotonic throughout the interval $(\mu_0,\mu_1)$, which preserves the ordering of samples between the two class means. This property implies that any threshold chosen within this interval therefore yields a consistent separation. We therefore adopt their midpoint $\tau$ as a symmetric and parameter-free central reference threshold. 

Outside the central monotonic interval, we introduce a second threshold $\tau_{\mathrm{side}}$, the reflection of $\tau$ about $d^\star$, since $\ell(d)$ is a quadratic symmetric around $d^\star$. This construction yields two thresholds that attain the same posterior value, while explicitly capturing the side region associated with the stationary point. Formally, the two thresholds are defined as
\begin{equation}
\label{eq:two_thresholds}
\tau := \frac{\mu_0+\mu_1}{2},
\qquad
\tau_{\mathrm{side}} := 2\,d^\star - \tau.
\end{equation}

\begin{figure}
    \centering
    \includegraphics[width=0.9\linewidth]{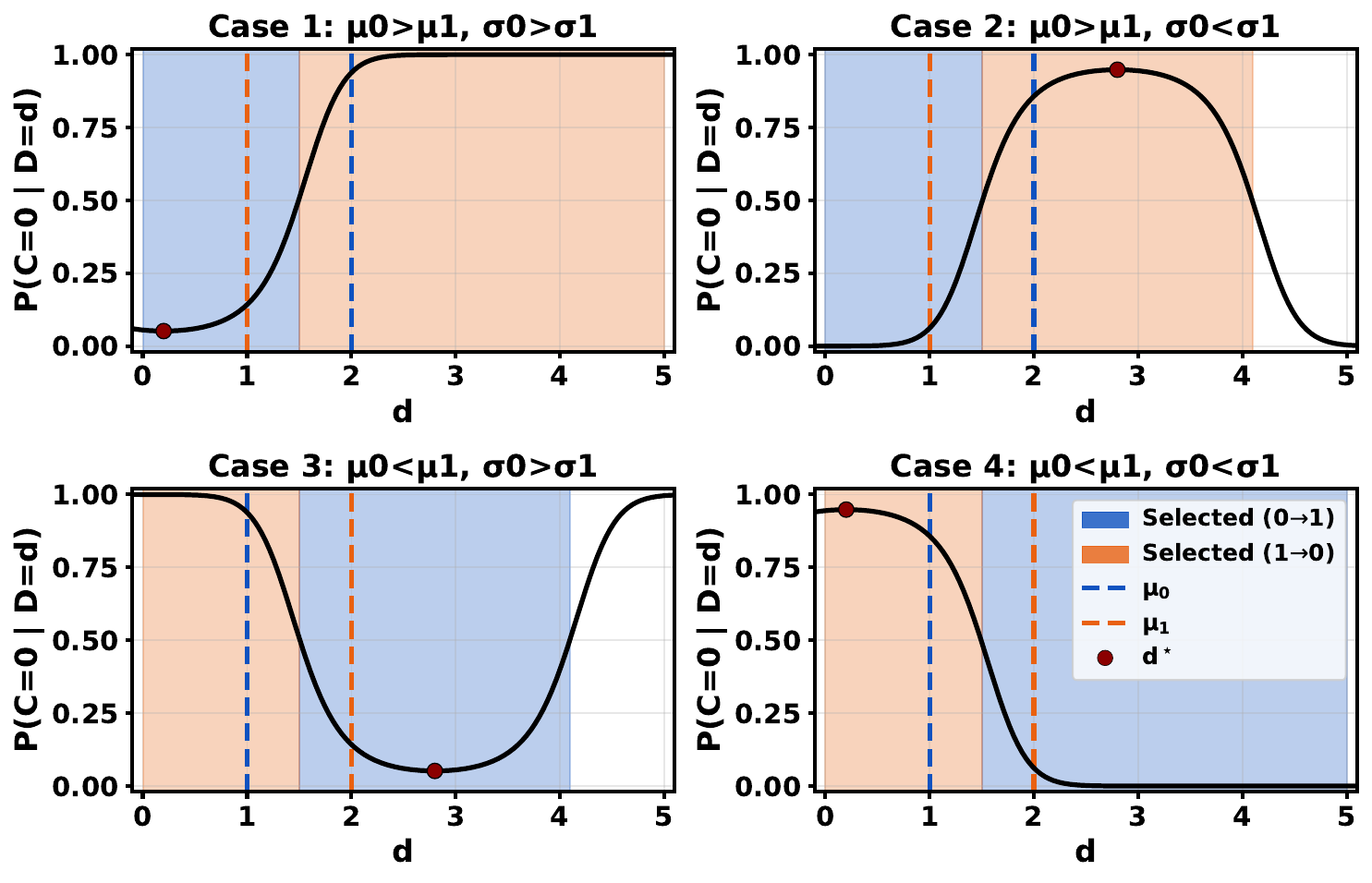}
    \caption{Posterior $\mathbb{P}(C=0\mid D=d)$ under four configurations of class means and standard deviations. Dashed lines denote the class means, the red dot indicates the stationary point, and shaded regions show samples selected by the proposed exchange rule.}
    \label{fig:cases}
\end{figure}

\subsection{Exchange Rule}
\label{sec:exchange}

With the thresholds $\tau$ and $\tau_{\mathrm{side}}$ defined in Eq.~\ref{eq:two_thresholds}, we now specify how posterior structure induces constraint exchange between offline and online behaviors. As illustrated in Fig.~\ref{fig:cases}, the posterior in Eq.~\ref{eq:posterior} exhibits four distinct qualitative patterns, depending on the relative ordering of the class means $(\mu_0,\mu_1)$ and variances $(\sigma_0,\sigma_1)$.

To capture the orientation of the posterior in a compact form, we introduce the direction signs
\begin{equation}
    \delta_\mu := \operatorname{sign}(\mu_0-\mu_1),
    \qquad
    \delta_\sigma := \operatorname{sign}(\sigma_0-\sigma_1).
    \label{eq:sign}
\end{equation}

Here, $\delta_\mu$ encodes the ordering of the class means, while $\delta_\sigma$ captures the direction of variance-induced curvature in the posterior.

Using the sign variables in Eq.~\ref{eq:sign}, we express the exchange rule in a unified form as follows:
\begin{mdframed}[linewidth=0.8pt,roundcorner=3pt]
\label{rule}
\noindent
An element with projection value $d$ from class $C=0$ is exchanged to class $C=1$ if and only if
\[
\delta_\mu\,\delta_\sigma\,(d-\tau_{\mathrm{side}})>0
\quad\text{and}\quad
\delta_\mu\,(d-\tau)<0.
\]
An element with projection value $d$ from class $C=1$ is exchanged to class $C=0$ if and only if
\[
\delta_\mu\,\delta_\sigma\,(d-\tau_{\mathrm{side}})>0
\quad\text{and}\quad
\delta_\mu\,(d-\tau)>0.
\]
\end{mdframed}

The two conditions in the exchange rule play complementary roles. The term \(\delta_\mu(d-\tau)\) determines on which side of the mean midpoint a sample lies, indicating whether it is closer to the opposite behavioral class. The term \(\delta_\mu\delta_\sigma(d-\tau_{\mathrm{side}})\) identifies the variance-dominated side region induced by the stationary point \(d^\star\), where the posterior curve deviates from simple mean-based separation. Exchanges are triggered only when both conditions are satisfied, ensuring that constraint reassignment occurs exclusively for samples that are both behaviorally ambiguous and located in the side region of the posterior curve. Since the rule is evaluated with support \(d\in\mathbb{R}_{\ge 0}\), any induced region with \(d<0\) is inactive and cannot trigger exchanges.

\begin{proposition}
\label{prop:four-case-exchange}
The sign-based exchange rule is equivalent to the four case-by-case conditions depicted in Fig.~\ref{fig:cases}.
\end{proposition}

\subsection{Exchange Operation}
\label{sec:operator}
The exchange rule in Sec.~\ref{sec:exchange} determines the candidate samples that should be reassigned between the two constraint roles. We then specify the corresponding exchange operation by separating the estimation of buffer-level exchange parameters from the sample-wise reassignment decision.

\textit{Buffer Level Statistics.~~}The quantities required by the rule, including \(\delta_\mu\), \(\delta_\sigma\), \(\tau\), and \(\tau_{\mathrm{side}}\), are induced by the buffer-level statistics \((\mu_0,\mu_1,\sigma_0,\sigma_1)\) through Eqs.~\ref{eq:d}--\ref{eq:sign}. This tuple is periodically updated from a reference set sampled from the offline and online buffers, allowing the exchange parameters to reflect the buffer-level behavior-distance distribution.

\textit{Mini-batch Level Operation.~~}Given these buffer-level exchange parameters, exchange is applied at the mini-batch level. Let \(b_{\mathrm{off}}\) and \(b_{\mathrm{on}}\) denote the offline and online mini-batches, respectively. The exchange rule in Sec.~\ref{sec:exchange} is then used to identify two candidate pools: an offline-to-online pool \(\mathcal{P}_{\mathrm{off} \rightarrow \mathrm{on}}\) and an online-to-offline pool \(\mathcal{P}_{\mathrm{on} \rightarrow \mathrm{off}}\). To keep the exchange operation balanced, we swap the same number of samples in the two directions
\[
K = \min\left(
|\mathcal{P}_{\mathrm{off}\rightarrow\mathrm{on}}|,
|\mathcal{P}_{\mathrm{on}\rightarrow\mathrm{off}}|
\right).
\]
When one candidate pool contains more than \(K\) samples, only \(K\) samples are selected from that pool, either randomly or according to the ordering of the behavioral distance \(d\). Notably, no additional exchange criterion is introduced at this stage, because samples in the candidate pools have already been identified by the exchange rule.

\subsection{Theoretical Analysis}
We analyze how exchanging samples between classes affects first-order statistics and distributional separability as characterized by the exchange rule.

Let $\mathcal{B}_0:=\{d^{(0)}_1,\dots,d^{(0)}_{n_0}\}$ and $\mathcal{B}_1:=\{d^{(1)}_1,\dots,d^{(1)}_{n_1}\}$ be two finite sets of projected values with sizes $n_0,n_1$. Select $x\in\mathcal{B}_0$ and $y\in\mathcal{B}_1$ such that
\(
\delta_\mu(x-\tau)<0
\,\text{and}\,
\delta_\mu(y-\tau)>0
\)
for exchange between the two sets. 

\begin{theorem}
\label{thm:mean-gap-increase}
Let $\mu_0',\mu_1'$ denote the empirical means after the exchange. Each exchange strictly increases the absolute mean gap
\[
|\Delta'|= |\mu_0'-\mu_1'| > |\Delta| = |\mu_0-\mu_1|.
\]
\end{theorem}

Thm.~\ref{thm:mean-gap-increase} establishes that the exchange procedure is mean-separation enhancing at the first-order level. To understand how the exchange reshapes the class-conditional distributions, we first characterize its effect on the empirical cumulative distribution functions (CDFs).

\begin{lemma}
\label{lem:swap-cdf}
Consider the empirical CDF
\[
F_c(t)=\frac{1}{n_c}\sum_{i=1}^{n_c}\mathbf{1}\{d^{(c)}_i\le t\},
\qquad c\in\{0,1\}.
\]
Suppose a single element with value $x$ in class $c$ is swapped with an element with value $y$ from the opposite class.
Let $F_c'$ denote the resulting empirical CDF after the swap.
Then, for every $t\in\mathbb{R}$,
\[
F_c'(t)
=
F_c(t)
- (-1)^c\,\delta_\mu\,\frac{1}{n_c}\,
\mathbf{1}\!\left\{
\delta_\mu\,x \le \delta_\mu\,t < \delta_\mu\,y
\right\}.
\]
\end{lemma}

We next consider the $H\Delta H$-divergence introduced by \citet{ben2010theory}, instantiated with the one-dimensional threshold hypothesis class
\[
H_{\mathrm{thr}}
=
\bigl\{ h_t(z)=\mathbf{1}\{z\le t\} : t\in\mathbb{R} \bigr\}.
\]
For two distributions $P$ and $Q$ over $Z\in\mathbb{R}$, the $H_{\mathrm{thr}}\Delta H_{\mathrm{thr}}$-divergence is defined as
\(
d_{H_{\mathrm{thr}}\Delta H_{\mathrm{thr}}}(P,Q).
\)

\begin{corollary}
\label{cor:cdf-hdh}
Define
\(
\Delta F(t) := F_0(t) - F_1(t),
\;
\Delta F_{\max} := \sup_{t\in\mathbb{R}} \Delta F(t),
\;
\Delta F_{\min} := \inf_{t\in\mathbb{R}} \Delta F(t),
\)
then
\[
d_{H_{\mathrm{thr}}\Delta H_{\mathrm{thr}}}(P,Q)
=
2\bigl(\Delta F_{\max}-\Delta F_{\min}\bigr).
\]
\end{corollary}
Combining the CDF perturbation behavior in Lem.~\ref{lem:swap-cdf} with the divergence characterization in Cor.~\ref{cor:cdf-hdh}, we obtain the following theorem.

\begin{theorem}
\label{lem:monotone-swap}
The exchange rule  does not decrease the $H\Delta H$-divergence for the threshold hypothesis class
\[
d_{H_{\mathrm{thr}}\Delta H_{\mathrm{thr}}}^{\;\prime}(P',Q')
\;\ge\;
d_{H_{\mathrm{thr}}\Delta H_{\mathrm{thr}}}(P,Q),
\]
where $d_{H_{\mathrm{thr}}\Delta H_{\mathrm{thr}}}(P,Q)$ and $d_{H_{\mathrm{thr}}\Delta H_{\mathrm{thr}}}^{\;\prime}(P',Q')$ denote the divergence before and after the swap, respectively.
\end{theorem}

Thms.~\ref{thm:mean-gap-increase} and~\ref{lem:monotone-swap} jointly establish that the exchange mechanism increases both mean separation and threshold-based distributional divergence, which provides a Bayes-optimal posterior guarantee for offline--online behavioral separability.

\subsection{Algorithm Summary}
We represent the learning agent as a collection of trainable components $\mathcal{C} = \{c_1, \dots, c_K\}$, which may include value functions, policies, or auxiliary networks depending on the underlying algorithm. The exchange mechanism operates on empirical statistics of alignment measure in Eq.~\ref{eq:distance}, specifically the mean $\mu$ and standard deviation $\sigma$, estimated periodically from both offline and online buffers. Algorithm~\ref{algo} summarizes the overall DARE procedure.
\begin{algorithm}[H]
\caption{Dynamic Alignment for RElaxation (DARE)}
\label{algo}
\begin{algorithmic}[1]
    \STATE \textbf{Initialize:} Offline components $\mathcal{C}$, behavior model $\pi_b$.
    \STATE \textbf{Initialize:} Offline dataset $\mathcal{D}_{\mathrm{off}}$, online buffer $\mathcal{D}_{\mathrm{on}}$.
    \FOR{each iteration}
        \STATE Collect online samples and store them in $\mathcal{D}_{\mathrm{on}}$.
        \STATE Sample a mini-batch $\{b_{\mathrm{off}},\ b_{\mathrm{on}}\}$ from $\mathcal{D}_{\mathrm{off}} $ and $ \mathcal{D}_{\mathrm{on}}$.
        \STATE Compute sample-wise alignment distance by Eq.~\ref{eq:distance}.
        \STATE Apply exchange rule to obtain exchange candidates.
        \STATE Apply exchange operation to obtain $\{b'_{\mathrm{off}},\ b'_{\mathrm{on}}\}$.
        \STATE Update $\mathcal{C}$ using behavior-specific objectives.
    \ENDFOR
\end{algorithmic}
\end{algorithm}

\section{DARE Instantiations}
The core of DARE is the exchange mechanism, which relies only on a per-sample alignment distance to a behavior model. Hence, it is compatible with a wide range of behavior models and fine-tuning objectives. We present representative instantiations for concreteness and empirical evaluation. Instantiation details are provided in \textbf{Appendix~\ref{appx:insta}}.

\textbf{Behavior Model.}
We primarily adopt an energy-guided diffusion model~\citep{lu2023contrastive,liu2024energy,zhang2024object,xu2025energy} trained on the offline dataset to produce offline-consistent reference actions. As an alternative instantiation, we also consider the online policy actor for online alignment, which constitutes a general and interchangeable instantiation of the behavioral reference.

\textbf{DARE in Cal-QL~} We extend Cal-QL with DARE, namely the variant DARE-C. During online fine-tuning, the conservative constraint is enforced only on offline-like samples to preserve stability and removed for online-like samples to enable adaptation.

\textbf{DARE in IQL.}
DARE-I preserves the original IQL value targets and updates the policy using the PEX objective~\citep{zhang2023policy} for offline-like samples. As for online-like samples, it replaces the value targets with TD-based maximum-Q estimates and adopts an entropy-regularized SAC objective~\citep{haarnoja2018soft} for policy. 

\begin{table*}[ht]
\centering
\resizebox{0.98\linewidth}{!}{\begin{threeparttable}
\caption{Performance after 0.2M online fine-tuning. Each result is averaged over the final 4 evaluations and 5 random seeds $\pm$ standard deviation. The ``-C'' and ``-I'' suffixes indicate the implementation based on Cal-QL and IQL, respectively. The highest scores are \textbf{bolded}.}
\label{tab:performance}
\begin{tabular}{l|cccc|cccc}
\toprule
\textbf{Dataset} & \multicolumn{4}{c}{\textbf{CQL Group}} & \multicolumn{4}{c}{\textbf{IQL Group}} \\
& Base (CQL) & Cal-QL & EDIS-C & DARE-C & Base (IQL)  & PEX & EDIS-I & DARE-I \\
\midrule
HC-ME  & 96.3\footnotesize{\( \pm \)1.6} & 96.4\footnotesize{\( \pm \)0.9} & 95.1\footnotesize{\( \pm \)0.7} & \textbf{97.1\footnotesize{\( \pm \)0.6}} 
       & 91.7\footnotesize{\( \pm \)2.3} & 89.2\footnotesize{\( \pm \)3.6} & 91.4\footnotesize{\( \pm \)3.9} & \textbf{93.6\footnotesize{\( \pm \)1.2}} \\
H-ME   & 111.9\footnotesize{\( \pm \)0.9} & 111.9\footnotesize{\( \pm \)0.7} & 111.9\footnotesize{\( \pm \)1.7} & 111.9\footnotesize{\( \pm \)0.4} 
       & 53.9\footnotesize{\( \pm \)39.0} & 90.2\footnotesize{\( \pm \)20.1} & 99.7\footnotesize{\( \pm \)13.5} & \textbf{104.5\footnotesize{\( \pm \)7.8}} \\
W2D-ME & 110.3\footnotesize{\( \pm \)0.5} & 110.4\footnotesize{\( \pm \)0.5} & 108.2\footnotesize{\( \pm \)7.1} & \textbf{112.3\footnotesize{\( \pm \)1.3}} 
       & 111.8\footnotesize{\( \pm \)4.9} & \textbf{114.8\footnotesize{\( \pm \)3.0}} & 113.1\footnotesize{\( \pm \)0.9} & \textbf{114.9\footnotesize{\( \pm \)2.1}} \\
HC-MR  & 50.9\footnotesize{\( \pm \)0.5} & 51.1\footnotesize{\( \pm \)1.1} & 56.4\footnotesize{\( \pm \)2.8} & \textbf{77.6\footnotesize{\( \pm \)2.6}} 
       & 47.4\footnotesize{\( \pm \)1.0} & \textbf{53.3\footnotesize{\( \pm \)1.2}} & 45.7\footnotesize{\( \pm \)0.7} & 48.4\footnotesize{\( \pm \)0.7} \\
H-MR   & 82.1\footnotesize{\( \pm \)33.2} & 93.0\footnotesize{\( \pm \)13.4} & 100.9\footnotesize{\( \pm \)5.9} & \textbf{102.7\footnotesize{\( \pm \)1.1}} 
       & 87.0\footnotesize{\( \pm \)28.2} & 93.5\footnotesize{\( \pm \)13.7} & 93.7\footnotesize{\( \pm \)9.7} & \textbf{104.2\footnotesize{\( \pm \)1.8}} \\
W2D-MR & 86.9\footnotesize{\( \pm \)3.4} & 88.4\footnotesize{\( \pm \)4.6} & 108.9\footnotesize{\( \pm \)4.2} & \textbf{110.2\footnotesize{\( \pm \)2.1}} 
       & 91.8\footnotesize{\( \pm \)6.2} & \textbf{92.0\footnotesize{\( \pm \)6.4}} & 89.2\footnotesize{\( \pm \)3.8} & 89.7\footnotesize{\( \pm \)3.2} \\
HC-M   & 64.6\footnotesize{\( \pm \)2.6} & 66.9\footnotesize{\( \pm \)1.8} & 68.8\footnotesize{\( \pm \)1.8} & \textbf{77.2\footnotesize{\( \pm \)3.4}} 
       & 57.8\footnotesize{\( \pm \)1.3} & 65.8\footnotesize{\( \pm \)2.9} & 49.3\footnotesize{\( \pm \)0.3} & \textbf{66.5\footnotesize{\( \pm \)1.4}} \\
H-M    & 81.9\footnotesize{\( \pm \)8.1} & 86.6\footnotesize{\( \pm \)9.1} & 94.7\footnotesize{\( \pm \)7.2} & \textbf{100.0\footnotesize{\( \pm \)0.8}} 
       & 77.5\footnotesize{\( \pm \)22.2} & 84.3\footnotesize{\( \pm \)20.1} & 58.9\footnotesize{\( \pm \)6.3} & \textbf{103.2\footnotesize{\( \pm \)1.5}} \\
W2D-M  & 83.0\footnotesize{\( \pm \)0.7} & 83.4\footnotesize{\( \pm \)1.5} & 85.9\footnotesize{\( \pm \)1.5} & \textbf{89.2\footnotesize{\( \pm \)3.2}} 
       & 85.8\footnotesize{\( \pm \)7.6} & 90.2\footnotesize{\( \pm \)13.1} & 86.4\footnotesize{\( \pm \)1.6} & \textbf{94.8\footnotesize{\( \pm \)4.1}} \\
\midrule
total (L) & 767.9 & 788.1 & 830.8 & \textbf{878.2} 
          & 704.7 & 773.3 & 727.4 & \textbf{819.8} \\
\midrule
AM-LD  & 53.8\footnotesize{\( \pm \)9.6} & 60.4\footnotesize{\( \pm \)7.7} & 70.1\footnotesize{\( \pm \)8.3} & \textbf{79.3\footnotesize{\( \pm \)8.9}} 
       & 50.6\footnotesize{\( \pm \)7.3} & 53.5\footnotesize{\( \pm \)9.5} & 52.8\footnotesize{\( \pm \)2.7} & \textbf{53.9\footnotesize{\( \pm \)5.3}} \\
AM-LP  & 50.2\footnotesize{\( \pm \)4.5} & 62.7\footnotesize{\( \pm \)9.5} & 68.2\footnotesize{\( \pm \)5.6} & \textbf{81.0\footnotesize{\( \pm \)6.0}} 
       & 45.5\footnotesize{\( \pm \)7.6} & 50.8\footnotesize{\( \pm \)10.2} & 48.2\footnotesize{\( \pm \)6.5} & \textbf{52.6\footnotesize{\( \pm \)4.2}} \\
AM-MD  & 85.8\footnotesize{\( \pm \)5.6} & 86.8\footnotesize{\( \pm \)4.7} & 93.4\footnotesize{\( \pm \)2.9} & \textbf{93.8\footnotesize{\( \pm \)3.4}} 
       & 82.0\footnotesize{\( \pm \)6.0} & 82.8\footnotesize{\( \pm \)6.4} & \textbf{84.9\footnotesize{\( \pm \)5.8}} & 81.1\footnotesize{\( \pm \)3.6} \\
AM-MP  & 86.2\footnotesize{\( \pm \)4.4} & 89.1\footnotesize{\( \pm \)5.4} & 94.4\footnotesize{\( \pm \)2.9} & 94.4\footnotesize{\( \pm \)5.2} 
       & 80.4\footnotesize{\( \pm \)3.6} & 81.7\footnotesize{\( \pm \)4.7} & 78.1\footnotesize{\( \pm \)7.2} & \textbf{83.1\footnotesize{\( \pm \)4.7}} \\
AM-UD  & 82.4\footnotesize{\( \pm \)7.4} & 90.1\footnotesize{\( \pm \)7.8} & 86.1\footnotesize{\( \pm \)3.7} & \textbf{91.3\footnotesize{\( \pm \)6.1}} 
       & 31.6\footnotesize{\( \pm \)16.1} & 30.6\footnotesize{\( \pm \)13.4} & 34.9\footnotesize{\( \pm \)9.3} & \textbf{69.2\footnotesize{\( \pm \)6.8}} \\
AM-U   & 93.2\footnotesize{\( \pm \)1.8} & 95.5\footnotesize{\( \pm \)1.3} & 95.2\footnotesize{\( \pm \)2.9} & \textbf{97.5\footnotesize{\( \pm \)2.9}} 
       & 90.5\footnotesize{\( \pm \)3.5} & 92.9\footnotesize{\( \pm \)3.7} & 92.2\footnotesize{\( \pm \)2.3} & \textbf{93.3\footnotesize{\( \pm \)2.7}} \\
\midrule
total (AM) & 451.6 & 484.6 & 507.4 & \textbf{537.3} 
           & 380.6 & 392.3 & 391.1 & \textbf{433.2} \\
\midrule
\midrule
total & 1219.5 & 1272.7 & 1338.2 & \textbf{1415.4} 
      & 1085.3 & 1165.6 & 1118.5 & \textbf{1253.0} \\
\bottomrule
\end{tabular}
\begin{tablenotes}
\footnotesize
\item[]\textit{Abbreviations:~} HC = \texttt{halfcheetah}, H = \texttt{hopper}, W2D = \texttt{walker2d}, L = \texttt{locomotion}, AM = \texttt{antmaze}; M = medium, ME = medium-expert, MR = medium-replay, LD = large-diverse, LP = large-play, MD = medium-diverse, MP = medium-play, UD = umaze-diverse, U = umaze. All environments use \texttt{-v2} version.
\end{tablenotes}
\end{threeparttable}}
\end{table*}

\section{Experiments}
We evaluate DARE and investigate the effectiveness, necessity, and robustness of the proposed exchange mechanism through the following key questions:
\begin{itemize}[itemsep=2pt, topsep=2pt, parsep=0pt]
    \item \textbf{Q1} Does DARE improve offline-to-online performance across benchmarks? (see in the Sec.~\ref{sec:overall})
    \item \textbf{Q2} Does the exchange mechanism remain active and adaptive throughout training? (see in the Sec.~\ref{sec:exchange_b})
    \item \textbf{Q3} Does a Bayes-optimal posterior play a necessary role in exchange decisions? (see in the Sec.~\ref{sec:bayes})
    \item \textbf{Q4} Does DARE generalize across different behavior models and online objectives? (see in the Sec.~\ref{sec:instantiation})
    \item \textbf{Q5} Does violating behavior-consistent exchange degrade learning performance? (see in the Sec.~\ref{sec:curiosity})
    \item \textbf{Q6} Does removing the exchange mechanism impair effective learning? (see in the Sec.~\ref{sec:ablation})
\end{itemize}

\subsection{Baselines and Benchmarks}
For baselines, we focus on methods that $(i)$ can be \emph{fully initialized} from the same offline-trained backbones and then fine-tuned online, and $(ii)$ differ primarily in \emph{constraint handling} rather than additional engineering choices (e.g., UTD, Q-ensemble). Accordingly, we group baselines by their offline initialization: in the \textbf{CQL group}, we compare DARE-C against CQL~\citep{kumar2020conservative}, Cal-QL, and EDIS-C~\citep{liu2024energy} (a Cal-QL variant); in the \textbf{IQL group}, we compare DARE-I against IQL, PEX~\citep{zhang2023policy}, and EDIS-I, the IQL counterpart of EDIS-C.

We evaluate DARE on three standard D4RL benchmarks~\citep{fu2020d4rl}: MuJoCo Locomotion, AntMaze Navigation, and Adroit Manipulation.
For fair comparison, EDIS-C and DARE-C in the CQL group are initialized from the same Cal-QL models. Meanwhile, all methods in the IQL group are initialized from the same IQL models. Additional implementation details, including the performance of the initial offline-trained models, are provided in \textbf{Appendix~\ref{appx:impl}}.

\subsection{Overall Performance}
\label{sec:overall}
The fine-tuning results are presented under both CQL group and IQL group in Tab.~\ref{tab:performance}. Overall, DARE brings a significant improvement of around $11\%$ total score of all datasets in both groups. Moreover, it consistently outperforms the baselines across different tasks, with highest scores on 13/15 tasks under the CQL group and on 12/15 tasks under the IQL group. These results demonstrate that the proposed exchange mechanism brings consistent benefits across diverse benchmarks in the O2O setting. \textbf{Appendix~\ref{appx:main}} provides the fine-tuning learning curves, illustrating that DARE maintains stable improvements across locomotion tasks. In addition, \textbf{Appendix~\ref{appx:additional-results}} reports additional results on Kitchen tasks, different instantiations of DARE-I, and comparisons using the TD3+BC backbone~\citep{fujimoto2021minimalist}. Together, these results validate the effectiveness of DARE's exchange mechanism across different experimental settings.

\begin{figure*}[ht]
    \centering
    \includegraphics[width=0.95\linewidth]{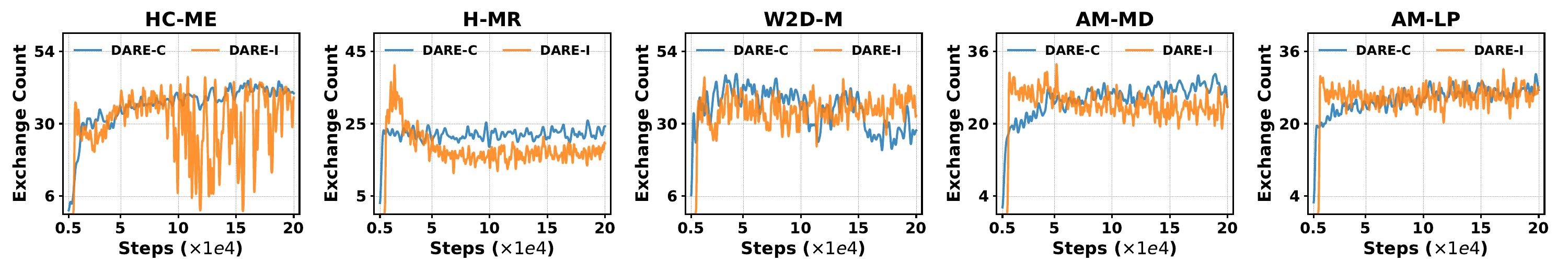}
    \caption{Adaptive exchange behavior of DARE-C and DARE-I during online training across different tasks.}
    \label{fig:exchange}
\end{figure*}
\begin{figure}[ht]
\centering
    \includegraphics[width=0.95\linewidth]{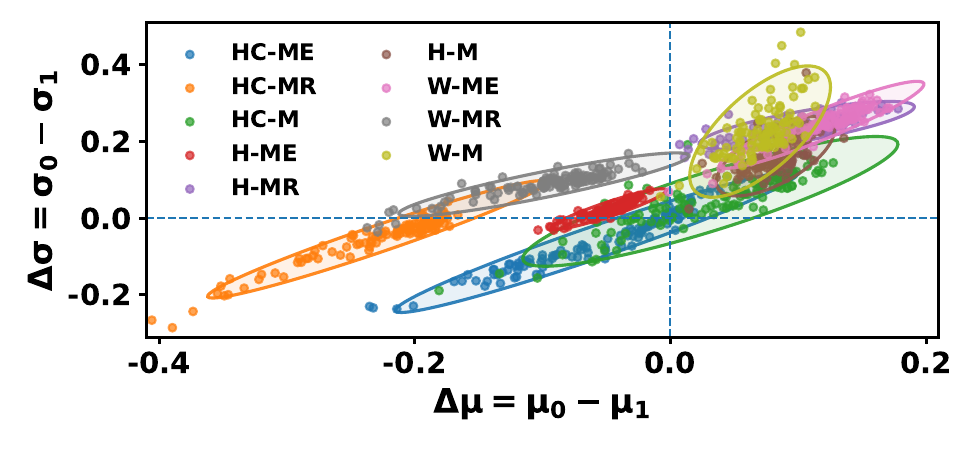}
    \caption{Estimated $\mu$ and $\sigma$ gaps during $0.2M$ training steps, both computed using a diffusion model trained on the offline dataset.}
    \label{fig:statistics}
\end{figure}
\subsection{Visualization of Exchange Behavior}
\label{sec:exchange_b}
DARE applies the exchange mechanism to reassign samples between offline and online batches. As shown in Fig.~\ref{fig:exchange}, the exchange pattern differs across environments, with the number of exchanges in most cases falling in the range of 20 to 50, which confirms that the proposed exchange mechanism actively performs sample reassignment during training. \textbf{Appendix~\ref{appx:exchange}} further reports the exchange pattern for all environments. We further depict the evolution of the estimated Gaussian statistics used in the Bayes-optimal formulation, as shown in Fig.~\ref{fig:statistics}. A natural intuition suggests that the online buffer should induce a larger mean than the offline dataset. However, the empirical estimates do not consistently follow the expected mean-based ordering across environments or training stages. This observation indicates that the exchange decision cannot rely on a simple monotonic separation between offline and online statistics.

\subsection{Effectiveness of Bayes-optimal Posterior}
\label{sec:bayes}
\begin{figure}[h]
    \centering
    \includegraphics[width=\linewidth]{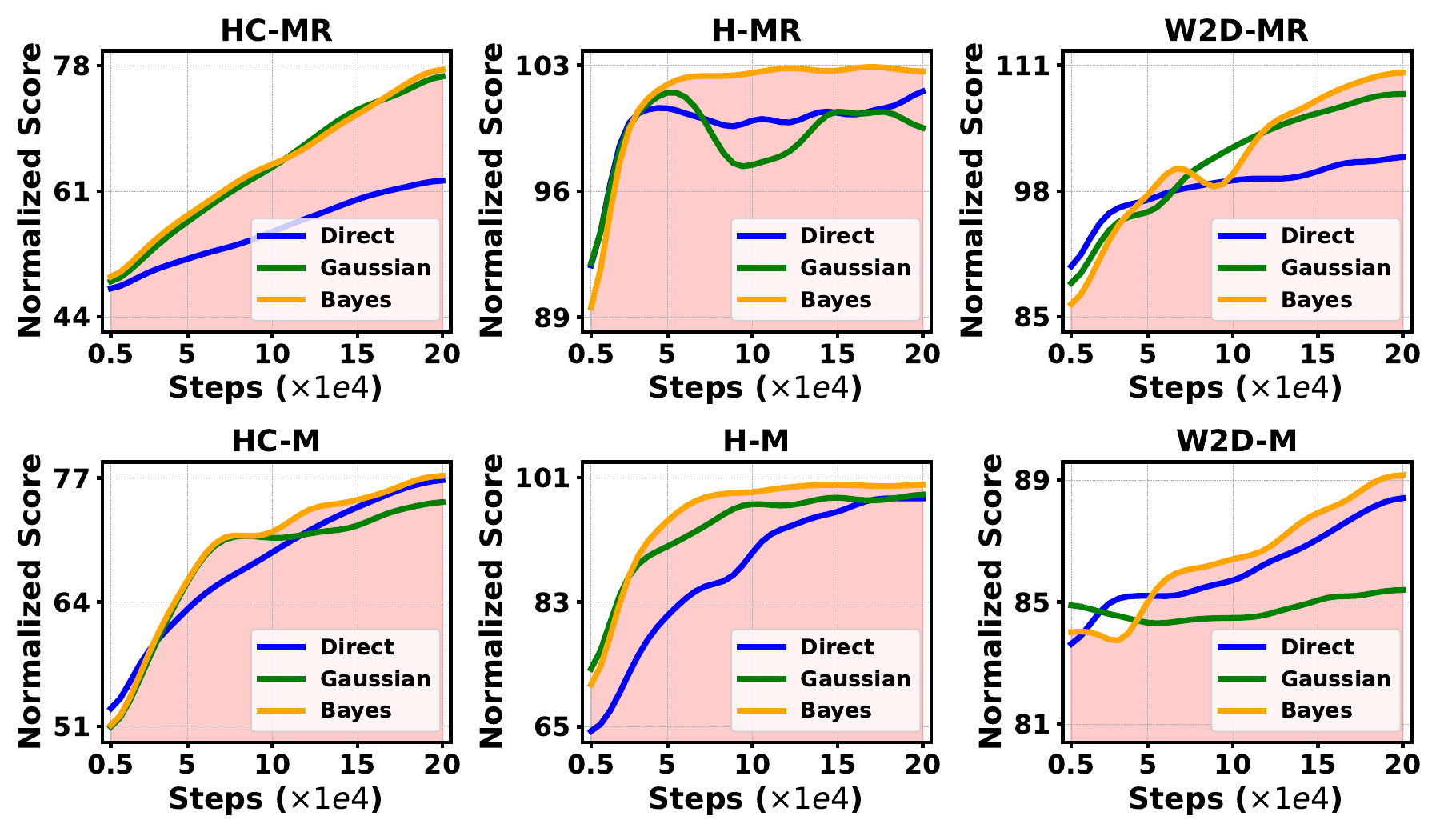}
    \caption{Comparison of posterior-induced exchange with monotonic Gaussian and direct alignment-distance on DARE-C.}
    \label{fig:cql_direct_fit}
\end{figure}
We compare three exchange mechanisms based on the same alignment measure in Eq.~\ref{eq:distance}$:$  (i) our posterior-induced exchange mechanism using the posterior $\mathbb{P}(C\,|\,D)$; (ii) a monotonic Gaussian exchange based on fitted distributions; and (iii) direct alignment-distance exchange via sample-level sorting. Fig.~\ref{fig:cql_direct_fit} shows that the posterior-induced exchange achieves more stable performance than the two monotonic strategies. This result also aligns with Fig.~\ref{fig:statistics}, which do not exhibit a consistent monotonic ordering. As a consequence, monotonic exchange mechanisms become unreliable, whereas the Bayes-optimal formulation remains robust on the full posterior. Importantly, the exchange uses the Bayes--optimal posterior to relax constraints only for high-confidence samples, thereby preventing extrapolation-driven updates in uncertain regions.

\subsection{Alternative Instantiations in DARE}
\label{sec:instantiation}
We evaluate DARE on the Adroit manipulation suite to test the generality of the exchange mechanism under different online-objective instantiations. In these experiments, we keep the exchange mechanism fixed but only replace the online objective applied to online-like samples. As summarized in Tab.~\ref{tab:adroit}, although the absolute performance varies across objective instantiations, the exchange mechanism yields consistent improvements across Adroit tasks. A detailed analysis of these results is provided in \textbf{Appendix~\ref{appx:adroit}}.
\begin{table}[H]
\centering
\caption{Adroit results for different instantiations of DARE-I. w/o S (no SAC), w/o P (no PEX), w/o S\&P (TD-only), with w/o E. and w/ E. indicating without and with exchange.}
\label{tab:adroit}
\resizebox{0.95\linewidth}{!}{
\renewcommand{\arraystretch}{1.05}
\begin{tabular}{l|cc|cc|cc}
\toprule
\textbf{Adroit}
& \multicolumn{2}{c|}{DARE-I.w/o S}
& \multicolumn{2}{c|}{DARE-I.w/o P}
& \multicolumn{2}{c}{DARE-I.w/o S\&P} \\
\texttt{human-v0}
& w/o E. & w/ E.
& w/o E. & w/ E.
& w/o E. & w/ E. \\
\midrule
Door     & \textbf{-0.02} & -0.15 & -0.07 & \textbf{31.36} & 17.31 & \textbf{19.90 }\\
Hammer   &  0.18 &  \textbf{0.26} & -0.04 &  \textbf{0.15} &  1.72 &  \textbf{4.04} \\
Pen      &106.35 &\textbf{113.48} &  9.17 & \textbf{23.55} & 85.10 & \textbf{97.82} \\
Relocate &  0.15 &  \textbf{0.27} & -0.10 &  \textbf{0.96} &  0.12 &  \textbf{0.39} \\
\midrule
\textbf{Total}    &106.67 &\textbf{113.90} &  8.96 & \textbf{56.02} &104.25 &\textbf{122.15} \\
\bottomrule
\end{tabular}}
\end{table}
We further consider an alternative instantiation of the behavior model in DARE. Instead of using a diffusion model trained on the offline dataset, we use the \texttt{online policy actor} as a consistency model for online-side behavior, reported in \textbf{Appendix~\ref{appx:behavior}}. This instantiation shows slightly lower performance than the diffusion-based offline behavior model. However, it avoids the computational cost of the reverse diffusion process and remains a practical option under limited computational budgets.

\subsection{Reversed Behavior Assignment}
\label{sec:curiosity}
\begin{table}[ht]
\centering
\caption{Results for reversed behavior assignment. Diffusion: offline diffusion as the behavior model, Policy: online policy actor as the behavior model, and DARE-I.C: curiosity-style DARE-I.}
\label{tab:curiosity}
\resizebox{0.85\linewidth}{!}{
\renewcommand{\arraystretch}{1.05}
\begin{tabular}{l|cc|cc}
\toprule
\textbf{Locomotion}
& \multicolumn{2}{c|}{Diffusion}
& \multicolumn{2}{c}{Policy}\\
& DARE-I.C & DARE-I
& DARE-I.C & DARE-I \\
\midrule
ME   &306.2 &\textbf{313.0} &294.7 &\textbf{311.7} \\
MR   &236.8 &\textbf{242.3} &232.7 &\textbf{236.5}  \\
M    &248.1 &\textbf{264.5} &239.1 &\textbf{263.9} \\
\midrule
\textbf{Total}    &791.1 &\textbf{819.8} &766.5 &\textbf{ 812.1}  \\
\bottomrule
\end{tabular}}
\end{table}
DARE assumes that samples can be categorized by behavioral similarity and trained with behavior-specific objectives. To examine this assumption, we perform an experiment that reverses the assignment strategy: Identified online-like samples are reassigned to the offline batch, while offline-like samples are reassigned to the online batch. This configuration represents the opposite operation of DARE, corresponding to a curiosity-style assignment. We also impose an exchange limit constant $n=32$, which matches the typical number of exchanges observed for DARE in Fig.~\ref{fig:exchange}. The corresponding results illustrated in Table~\ref{tab:curiosity}, which supports behavioral consistency as the criterion for role assignment. More experimental results are provided in \textbf{Appendix~\ref{appx:reverse}}.

\subsection{Ablation Study}
\label{sec:ablation}
\begin{figure}[ht]
  \centering    \includegraphics[width=1\linewidth]{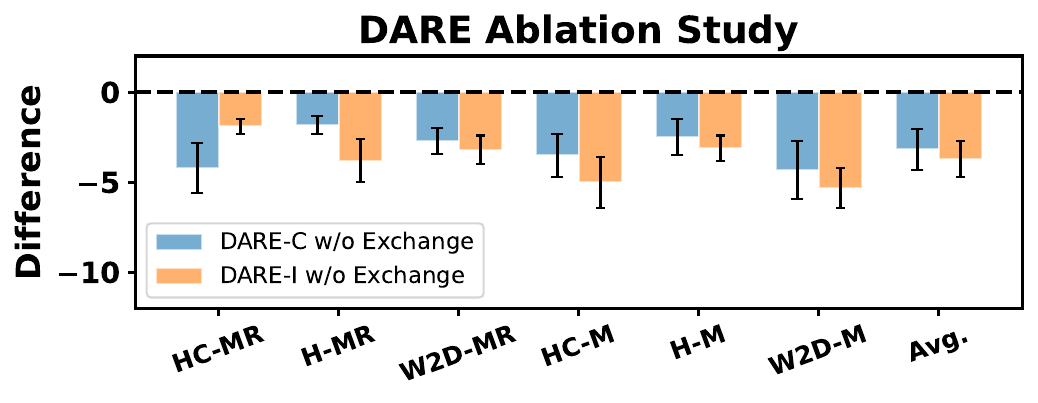}
    \caption{Ablation study of DARE.}
\label{fig:ablation}
\end{figure}
Finally, we conduct an ablation study that removes the exchange mechanism, as shown in Fig.~\ref{fig:ablation}. The figure compares DARE-C and DARE-I with their corresponding w/o Exchange variants across multiple environments and reports the average performance difference. In all evaluated environments, both DARE-C and DARE-I show positive differences relative to their ablated counterparts, which reflects performance degradation after removal of the exchange operation. Additional ablation results are provided in \textbf{Appendix~\ref{appx:abla}}, including the choice of behavioral distance measure $d$ in Eq.~\ref{eq:distance} and runtime comparisons.

\section{Limitations}
\label{sec:limitations}
Although DARE improves offline-to-online transfer across different settings, its effectiveness can still depend on the quality and coverage of the offline dataset. In DARE, the behavior model is first trained from offline data and then continuously updated during online fine-tuning. When the offline dataset has limited coverage or a substantial mismatch from the online trajectory, the behavior model may produce less reliable judgments, affecting the transition from offline regularization to online adaptation. In addition to this data-dependent limitation, our theory analysis focuses on the exchange rule rather than the full end-to-end O2O learning process. A complete theory linking exchange decisions to value estimation and policy improvement remains an important direction for future work.

\section{Conclusion}
We address a fundamental limitation in O2O RL: distribution shift during fine-tuning induces out-of-support behaviors that existing approaches fail to handle at the sample level. To this end, we propose Dynamic Alignment for RElaxation (DARE), a distribution-aware framework that performs sample-level constraint relaxation based on behavioral consistency via a posterior-induced exchange mechanism. Theoretically, we demonstrate that the exchange mechanism follows from a Bayes-optimal posterior formulation and provably enhances offline--online behavioral separability. Importantly, DARE is algorithm-agnostic, admitting instantiations across diverse behavior-model choices and O2O algorithms. Empirically, DARE consistently improves performance over the corresponding baselines across all instantiated variants on standard benchmarks. These findings suggest that DARE provides a theoretically grounded foundation for bridging the offline-to-online gap through behavior-consistent, sample-level constraint adaptation.

\section*{Acknowledgements}
We thank Hansong Zhou and Yukun Yuan for helpful discussions, and the anonymous reviewers for their constructive comments.

\section*{Impact Statement}
This paper advances the study of offline-to-online RL by addressing the mismatch between data distributions and learning objectives during transfer. There are many potential societal consequences of our work, none of which we feel must be specifically highlighted here.

\newpage

\bibliography{example_paper}
\bibliographystyle{icml2026}


\setcounter{theorem}{0}
\setcounter{proposition}{0}
\setcounter{lemma}{0}
\setcounter{corollary}{0}
\setcounter{property}{0}
\setcounter{definition}{0}
\setcounter{assumption}{0}
\setcounter{remark}{0}
\newpage
\appendix
\onecolumn

\section*{Appendix Guide}
This appendix provides the supporting details for the main paper, organized as follows:

\apxAauto{\ref{appx:proofs}}{Proofs}{appx:proofs}
\apxAoneauto{Thm.~\ref{appx:thm1}}{Sufficiency of \(d\) for Bayes-optimal Posterior Inference}{appx:thm1}
\apxAoneauto{Thm.~\ref{appx:thm2}}{Stationary Point Location for the Quadratic-logistic Posterior}{appx:thm2}
\apxAoneauto{Prop.~\ref{appx:prop1}}{Case-wise Form of the Sign-based Exchange Rule}{appx:prop1}
\apxAoneauto{Thm.~\ref{appx:thm3}}{Increase of the Absolute Mean Separation}{appx:thm3}
\apxAoneauto{Lemm.~\ref{appx:lemm1}}{Piecewise Characterization of Post-exchange CDFs}{appx:lemm1}
\apxAoneauto{Cor.~\ref{appx:cor1}}{Closed Form of \(H_{\mathrm{thr}}\Delta H_{\mathrm{thr}}\) via CDFs}{appx:cor1}
\apxAoneauto{Thm.~\ref{appx:thm4}}{Non-decreasing Threshold Separation under Exchange}{appx:thm4}
\apxAauto{\ref{appx:gauss}}{Analysis of the Gaussian Assumption}{appx:gauss}
\apxAauto{\ref{appx:insta}}{DARE Instantiation Details}{appx:insta}
\apxAauto{\ref{appx:impl}}{Experimental Details}{appx:impl}
\apxAauto{\ref{appx:main}}{Comparisons of Online Fine-tuning Processes}{appx:main}
\apxAauto{\ref{appx:additional-results}}{Additional Results for Main Table}{appx:additional-results}
\apxAauto{\ref{appx:exchange}}{Visualization of Exchange Behavior}{appx:exchange}
\apxAauto{\ref{appx:adroit}}{Further Results and Analysis on Adroit}{appx:adroit}
\apxAauto{\ref{appx:behavior}}{Alternative Behavior Model Instantiations}{appx:behavior}
\apxAauto{\ref{appx:reverse}}{Additional Results for Reversed Behavior Assignment}{appx:reverse}
\apxAauto{\ref{appx:abla}}{Additional Ablation Study}{appx:abla}

\newpage
\section{Proofs}
\label{appx:proofs}
\begin{definition}
Let $(s, a)$ denote a state--action pair taking values in space $\mathcal{S}\times\mathcal{A}$, and let $\pi_b(\cdot\mid s)$ denote a reference behavior model. Define a one-dimensional behavioral deviation statistic
\[
d(s,a) := \operatorname{dist}\bigl(a, a_b(s)\bigr),\ \mathcal{S}\times\mathcal{A}\to\mathbb{R}, 
\]
where $\operatorname{dist}(\cdot,\cdot)$ is a fixed deviation measure and $a_b(s)$ denotes the reference action by $\pi_b$. 
\end{definition}

\begin{theorem}
\phantomsection
\label{appx:thm1}
Suppose the class-conditional densities admit a likelihood--ratio representation through \(d\); that is, there exists a measurable function
\(r:\mathbb{R}\to(0,\infty)\) such that
\[
\frac{p(X=x\mid C=0)}{p(X=x\mid C=1)} = r\!\bigl(d(s,a)\bigr).
\]
It follows that \(d\) is sufficient for Bayes--optimal posterior inference between the two classes, i.e. there exists a measurable function \(g:\mathbb{R}\to[0,1]\) such that
\[
\mathbb{P}(C=0 \mid X=x) = g\!\bigl(d(s, a)\bigr).
\]
\end{theorem}

\begin{proof}
Fix any state-conditioned action variable $x$ and by Bayes' rule, the posterior probability can be written as
\[
\mathbb{P}(C=0 \mid X=x)
=
\frac{\mathbb{P}(C=0)\,p_0(x)}
     {\mathbb{P}(C=0)\,p_0(x)+\mathbb{P}(C=1)\,p_1(x)}.
\]
Dividing the numerator and denominator by $p_1(x)$ yields
\[
\mathbb{P}(C=0 \mid X=x)
=
\frac{\mathbb{P}(C=0)\,\frac{p_0(x)}{p_1(x)}}
     {\mathbb{P}(C=0)\,\frac{p_0(x)}{p_1(x)}+\mathbb{P}(C=1)}
=
\frac{\mathbb{P}(C=0)\,r\!\bigl(d(s,a)\bigr)}
     {\mathbb{P}(C=0)\,r\!\bigl(d(s,a)\bigr)+\mathbb{P}(C=1)}.
\]
Since the right-hand side is a function of $r\!\bigl(d(s,a)\bigr)$ alone, there exists a measurable function $g:\mathbb{R}\to(0,1)$ such that
\[
\mathbb{P}(C=0 \mid X=x)=g(d(s,a)).
\]
\end{proof}

\begin{property}
\label{prop:second-order}
Let $D:\mathcal{S}\times\mathcal{A}\to\mathbb{R}$ be a real-valued random variable and let $C\in\{0,1\}$ denote a binary class label. We say that the posterior distribution of $C$ given $D$ is \emph{second-order Gaussian-determined} if there exist parameters $\mu_0,\mu_1\in\mathbb{R}$ and $\sigma_0^2,\sigma_1^2>0$ such that
\[
D \mid C=c \sim \mathcal{N}(\mu_c,\sigma_c^2),
\qquad c\in\{0,1\}.
\]
Under this model, the class-conditional densities of $D$ are given by
\[
p(D=d \mid C=0)
=
\frac{1}{\sqrt{2\pi}\,\sigma_0}
\exp\!\left(-\frac{(d-\mu_0)^2}{2\sigma_0^2}\right),
\qquad
p(D=d \mid C=1)
=
\frac{1}{\sqrt{2\pi}\,\sigma_1}
\exp\!\left(-\frac{(d-\mu_1)^2}{2\sigma_1^2}\right).
\]
\end{property}

\begin{property}
\label{prop:posterior}
Let $\rho_c := \mathbb{P}(C=c)$ with $\rho_0,\rho_1>0$ and $\rho_0+\rho_1=1$. Under this model, the posterior probability $\mathbb{P}(C=0 \mid D=d)$. Then, for any $d\in\mathbb{R}$, the posterior probability is
\[
\mathbb{P}(C=0 \mid D=d)
=
\frac{\rho_0\,\sigma_1
\exp\!\left(-\frac{(d-\mu_0)^2}{2\sigma_0^2}\right)}
{\rho_0\,\sigma_1
\exp\!\left(-\frac{(d-\mu_0)^2}{2\sigma_0^2}\right)
+
\rho_1\,\sigma_0
\exp\!\left(-\frac{(d-\mu_1)^2}{2\sigma_1^2}\right)}
=
\frac{1}{1+\frac{\rho_1\,\sigma_0}{\rho_0\,\sigma_1}\exp\!\left(-\frac{(d-\mu_1)^2}{2\sigma_1^2}+\frac{(d-\mu_0)^2}{2\sigma_0^2}\right)}
\]
\end{property}

\begin{property}
\label{prop:logistic-quadratic}
Let $\alpha>0$ and define
\[
f(x):=\frac{1}{1+\alpha\,\exp(-x)}.
\]
Then $f$ is strictly increasing in $x$, with derivative
\[
f'(x)=\frac{\alpha\,\exp(-x)}{\bigl(1+\alpha\,\exp(-x)\bigr)^2}>0.
\]
Under Property~\ref{prop:posterior}, define
\[
x: \ell(d)=\frac{(d-\mu_1)^2}{2\sigma_1^2}-\frac{(d-\mu_0)^2}{2\sigma_0^2},
\qquad
\alpha:=\frac{\rho_1\,\sigma_0}{\rho_0\,\sigma_1}.
\]
Then the posterior can be written as
\[
\mathbb{P}(C=0\mid D=d)=f\bigl(\ell(d)\bigr).
\]
Moreover, $\ell(d)$ is a quadratic function of $d$:
\[
\ell(d)=a\,d^2+b\,d+c,
\]
where the quadratic coefficient is
\[
a=\frac{1}{2}\!\left(\frac{1}{\sigma_1^2}-\frac{1}{\sigma_0^2}\right),
\qquad
b=-\frac{\mu_1}{\sigma_1^2}+\frac{\mu_0}{\sigma_0^2},
\qquad
c=\frac{\mu_1^2}{2\sigma_1^2}-\frac{\mu_0^2}{2\sigma_0^2}.
\]

Consequently,
\[
x'(d)=2a\,d+b,
\]
and when $a\neq 0$ the unique stationary point (extremum) of $x(d)$ is
\[
d^\star=-\frac{b}{2a}
=
\frac{\mu_1/\sigma_1^2-\mu_0/\sigma_0^2}{1/\sigma_1^2-1/\sigma_0^2}.
\]
\end{property}

\begin{theorem}
\phantomsection
\label{appx:thm2}
The stationary point $d^\star$ in Property~\ref{prop:logistic-quadratic} satisfies $d^\star\notin(\mu_0,\mu_1)$.
\end{theorem}

\begin{proof}
From Property~\ref{prop:logistic-quadratic},
\[
d^\star=\frac{\mu_1\sigma_0^2-\mu_0\sigma_1^2}{\sigma_0^2-\sigma_1^2}
=\mu_0+(\mu_1-\mu_0)\frac{\sigma_0^2}{\sigma_0^2-\sigma_1^2}.
\]
Let $r:=\frac{\sigma_0^2}{\sigma_0^2-\sigma_1^2}$. Since $\sigma_0^2\neq\sigma_1^2$, either $r>1$ (when $\sigma_0^2>\sigma_1^2$) or $r<0$ (when $\sigma_0^2<\sigma_1^2$). We consider four cases:

\emph{Case 1:} $\mu_0<\mu_1$ and $\sigma_0^2>\sigma_1^2$. Then $\mu_1-\mu_0>0$ and $r>1$, hence
\[
d^\star=\mu_0+(\mu_1-\mu_0)r>\mu_0+(\mu_1-\mu_0)=\mu_1.
\]

\emph{Case 2:} $\mu_0<\mu_1$ and $\sigma_0^2<\sigma_1^2$. Then $\mu_1-\mu_0>0$ and $r<0$, hence
\[
d^\star=\mu_0+(\mu_1-\mu_0)r<\mu_0.
\]

\emph{Case 3:} $\mu_1<\mu_0$ and $\sigma_0^2>\sigma_1^2$. Then $\mu_1-\mu_0<0$ and $r>1$, hence
\[
d^\star=\mu_0+(\mu_1-\mu_0)r<\mu_0+(\mu_1-\mu_0)=\mu_1.
\]

\emph{Case 4:} $\mu_1<\mu_0$ and $\sigma_0^2<\sigma_1^2$. Then $\mu_1-\mu_0<0$ and $r<0$, hence
\[
d^\star=\mu_0+(\mu_1-\mu_0)r>\mu_0.
\]

In all cases, $d^\star$ lies outside the open interval with endpoints $\mu_0$ and $\mu_1$, i.e., $d^\star\notin(\min\{\mu_0,\mu_1\},\max\{\mu_0,\mu_1\})$. In particular, under $\mu_0<\mu_1$, we have $d^\star\notin(\mu_0,\mu_1)$.
\end{proof}

\begin{definition}
\label{def:exchange}
Let $D:\mathcal{S}\times\mathcal{A}\to\mathbb{R}$ be a real-valued random variable and let $C\in\{0,1\}$ denote a binary class label. Denote the class-wise means and standard deviations by $\mu_0, \mu_1, \sigma_0, \sigma_1$. Define the mean-direction and dispersion-direction signs as
\[
\delta_\mu := \operatorname{sign}(\mu_0-\mu_1),
\qquad
\delta_\sigma := \operatorname{sign}(\sigma_0-\sigma_1),
\]
and the thresholds
\[
\tau := \frac{\mu_0+\mu_1}{2},
\qquad
\tau_{\mathrm{side}} := 2\,d^\star-\tau,
\]
where $d^\star$ is stationary point.

An element with projection value $d$ belonging to class $C=0$
is selected for exchange to class $C=1$ if and only if
\[
\delta_\mu\,\delta_\sigma\,(d-\tau_{\mathrm{side}})>0
\quad\text{and}\quad
\delta_\mu\,(d-\tau)<0.
\]

An element with projection value $d$ belonging to class $C=1$
is selected for exchange to class $C=0$ if and only if
\[
\delta_\mu\,\delta_\sigma\,(d-\tau_{\mathrm{side}})>0
\quad\text{and}\quad
\delta_\mu\,(d-\tau)>0.
\]
\end{definition}

\begin{proposition}
\phantomsection
\label{appx:prop1}
The sign-based exchange rule in Definition~\ref{def:exchange} is equivalent to the following four case-by-case conditions:
\emph{Case 1:} If $\mu_0>\mu_1$ and $\sigma_0>\sigma_1$
(i.e., $\delta_\mu=+1,\ \delta_\sigma=+1$), then
\[
\text{(0$\to$1)}:\ \ \tau_{\mathrm{side}}<d<\tau,
\qquad
\text{(1$\to$0)}:\ \ d>\max\{\tau,\tau_{\mathrm{side}}\}.
\]
\emph{Case 2:} If $\mu_0>\mu_1$ and $\sigma_0<\sigma_1$
(i.e., $\delta_\mu=+1,\ \delta_\sigma=-1$), then
\[
\text{(0$\to$1)}:\ \ d<\min\{\tau,\tau_{\mathrm{side}}\},
\qquad
\text{(1$\to$0)}:\ \ \tau<d<\tau_{\mathrm{side}}.
\]
\emph{Case 3:} If $\mu_0<\mu_1$ and $\sigma_0>\sigma_1$
(i.e., $\delta_\mu=-1,\ \delta_\sigma=+1$), then
\[
\text{(0$\to$1)}:\ \ \tau<d<\tau_{\mathrm{side}},
\qquad
\text{(1$\to$0)}:\ \ d<\min\{\tau,\tau_{\mathrm{side}}\}.
\]
\emph{Case 4:} If $\mu_0<\mu_1$ and $\sigma_0<\sigma_1$
(i.e., $\delta_\mu=-1,\ \delta_\sigma=-1$), then
\[
\text{(0$\to$1)}:\ \ d>\max\{\tau,\tau_{\mathrm{side}}\},
\qquad
\text{(1$\to$0)}:\ \ \tau_{\mathrm{side}}<d<\tau.
\]
\end{proposition}

\begin{proof}
For class $0\to 1$, the rule requires simultaneously
\(
\delta_\mu\delta_\sigma(d-\tau_{\mathrm{side}})>0
\,\text{and}\,
\delta_\mu(d-\tau)<0.
\)
If $\delta_\mu=+1$, the second inequality is $d<\tau$; if $\delta_\mu=-1$, it is $d>\tau$. Likewise, if $\delta_\mu\delta_\sigma=+1$ the first inequality is $d>\tau_{\mathrm{side}}$, whereas if $\delta_\mu\delta_\sigma=-1$ it becomes $d<\tau_{\mathrm{side}}$. Intersecting these two half-spaces yields exactly the interval conditions listed for (0$\to$1) in each case.

The class $1\to 0$ rule is identical except that the second inequality flips to $\delta_\mu(d-\tau)>0$, which becomes $d>\tau$ when $\delta_\mu=+1$ and $d<\tau$ when $\delta_\mu=-1$. Combining with the same first inequality $\delta_\mu\delta_\sigma(d-\tau_{\mathrm{side}})>0$ yields the listed conditions for (1$\to$0) in each case.
\end{proof}

\begin{theorem}
\phantomsection
\label{appx:thm3}
Let $\mathcal{B}_0,\mathcal{B}_1$ be two finite sets of projected values with empirical means $\mu_0,\mu_1$ and sizes $n_0,n_1$. Define
\[
\Delta:=\mu_0-\mu_1,\qquad \delta_\mu:=\operatorname{sign}(\Delta),\qquad
\tau:=\frac{\mu_0+\mu_1}{2}.
\]
Select $x\in\mathcal{B}_0$ and $y\in\mathcal{B}_1$ such that
\[
\delta_\mu(x-\tau)<0
\quad\text{and}\quad
\delta_\mu(y-\tau)>0,
\]
and swap $x$ and $y$ between the two sets. Let $\mu_0',\mu_1'$ denote the empirical means after the exchange and define $\Delta':=\mu_0'-\mu_1'$. Each exchange strictly increases the absolute mean gap:
\[
|\Delta'|>|\Delta|.
\]
\end{theorem}

\begin{proof}
After swapping $x\in\mathcal{B}_0$ with $y\in\mathcal{B}_1$, the batch sizes remain $n_0,n_1$ and the new means satisfy
\[
\mu_0'=\mu_0+\frac{y-x}{n_0},
\qquad
\mu_1'=\mu_1+\frac{x-y}{n_1}.
\]
Hence the new mean gap is
\[
\Delta'=\mu_0'-\mu_1'
=
(\mu_0-\mu_1)+\Bigl(\frac{1}{n_0}+\frac{1}{n_1}\Bigr)(y-x)
=
\Delta+\kappa (y-x),
\]
where $\kappa:=\frac{1}{n_0}+\frac{1}{n_1}>0$.

It remains to show that $\delta_\mu (y-x)>0$. From the side conditions,
\[
\delta_\mu(x-\tau)<0 \ \Longleftrightarrow\ \delta_\mu x<\delta_\mu \tau,
\qquad
\delta_\mu(y-\tau)>0 \ \Longleftrightarrow\ \delta_\mu y>\delta_\mu \tau.
\]
Combining them yields
\[
\delta_\mu y>\delta_\mu \tau>\delta_\mu x
\quad\Longrightarrow\quad
\delta_\mu (y-x)>0.
\]
Therefore,
\[
\delta_\mu \Delta'
=
\delta_\mu \Delta+\kappa\,\delta_\mu (y-x)
=
|\Delta|+\kappa\,|y-x|
>
|\Delta|.
\]
Since $\delta_\mu\Delta'>0$, we also have $\operatorname{sign}(\Delta')=\delta_\mu=\operatorname{sign}(\Delta)$.

Finally,
\[
|\Delta'|
=
\delta_\mu \Delta'
=
|\Delta|+\kappa|y-x|
\quad\Longrightarrow\quad
|\Delta'|>|\Delta|.
\]
This proves the claim.
\end{proof}

\begin{lemma}
\phantomsection
\label{appx:lemm1}
Consider the empirical CDF
\[
F_c(t)=\frac{1}{n_c}\sum_{i=1}^{n_c}\mathbf{1}\{d^{(c)}_i\le t\},
\qquad c\in\{0,1\}.
\]
Suppose a single element with value $x$ in class $c$ is swapped with an element with value $y$ from the opposite class.
Let $F_c'$ denote the resulting empirical CDF after the swap.
Then, for every $t\in\mathbb{R}$,
\[
F_c'(t)
=
F_c(t)
- (-1)^c\,\delta_\mu\,\frac{1}{n_c}\,
\mathbf{1}\!\left\{
\delta_\mu\,x \le \delta_\mu\,t < \delta_\mu\,y
\right\}.
\]
\end{lemma}

\begin{proof}
By construction, all projected values remain unchanged except for the replacement of $x$ by $y$ in class $0$ and the replacement of $y$ by $x$ in class $1$. Hence for any $t$,
\[
F_0'(t)-F_0(t)
=
\frac{1}{n_0}\Bigl(\mathbf{1}\{y\le t\}-\mathbf{1}\{x\le t\}\Bigr),
\qquad
F_1'(t)-F_1(t)
=
\frac{1}{n_1}\Bigl(\mathbf{1}\{x\le t\}-\mathbf{1}\{y\le t\}\Bigr).
\]
It remains to simplify these indicator differences under the selection conditions. Since $\delta_\mu(x-\tau)<0$ and $\delta_\mu(y-\tau)>0$, we have
\[
\delta_\mu x<\delta_\mu\tau<\delta_\mu y
\quad\Longrightarrow\quad
\delta_\mu(y-x)>0,
\]
Consider the event
\[
I
=
\{\,\delta_\mu\,y \le \delta_\mu\,t < \delta_\mu\,x\,\}
=
\begin{cases}
\{t:\ y\le t<x\}, & \text{if }\delta_\mu=+1,\\[2pt]
\{t:\ x\le t<y\}, & \text{if }\delta_\mu=-1.
\end{cases}
\]
For $t\notin I$, either $t$ is on the same side of both $x$ and $y$ (so $\mathbf{1}\{y\le t\}=\mathbf{1}\{x\le t\}$), or $t$ exceeds both (so again the two indicators coincide). Thus $F_0'(t)=F_0(t)$ and $F_1'(t)=F_1(t)$ for $t\notin I$.

For $t\in I$, exactly one of $\{x\le t\}$ and $\{y\le t\}$ holds, and since $\delta_\mu(y-x)>0$ we have
\[
\mathbf{1}\{y\le t\}-\mathbf{1}\{x\le t\}=-\delta_\mu,
\qquad
\mathbf{1}\{x\le t\}-\mathbf{1}\{y\le t\}=\delta_\mu.
\]
Substituting into the CDF differences yields
\[
F_0'(t)=F_0(t)-\delta_\mu\,\frac{1}{n_0},\qquad
F_1'(t)=F_1(t)+\delta_\mu\,\frac{1}{n_1},
\qquad \forall t\in I,
\]
which is precisely
\[
F_0'(t)
=
F_0(t)
-\delta_\mu\,\frac{1}{n_0}\mathbf{1}\{t\in I\},
\qquad
F_1'(t)
=
F_1(t)
+\delta_\mu\,\frac{1}{n_1}\mathbf{1}\{t\in I\}.
\]
This completes the proof.
\end{proof}

\begin{definition}
We consider the $H\Delta H$-divergence introduced by \citet{ben2010theory}, instantiated with the one-dimensional threshold hypothesis class
\[
H_{\mathrm{thr}}
=
\bigl\{ h_t(z)=\mathbf{1}\{z\le t\} : t\in\mathbb{R} \bigr\}.
\]
For two distributions $P$ and $Q$ over $Z\in\mathbb{R}$, the $H_{\mathrm{thr}}\Delta H_{\mathrm{thr}}$-divergence is defined as
\[
d_{H_{\mathrm{thr}}\Delta H_{\mathrm{thr}}}(P,Q)
\;:=\;
2\sup_{h,h'\in H_{\mathrm{thr}}}
\bigl|P\big(h(Z)\ne h'(Z)\big)-Q\big(h(Z)\ne h'(Z)\big)\bigr|.
\]
\end{definition}

\begin{corollary}
\phantomsection
\label{appx:cor1}
Let $F_0$ and $F_1$ denote the cumulative distribution functions (CDFs) of $P$ and $Q$, respectively. Define
\[
\Delta F(t) := F_0(t) - F_1(t),
\qquad
\Delta F_{\max} := \sup_{t\in\mathbb{R}} \Delta F(t),
\qquad
\Delta F_{\min} := \inf_{t\in\mathbb{R}} \Delta F(t).
\]
Then
\[
d_{H_{\mathrm{thr}}\Delta H_{\mathrm{thr}}}(P,Q)
=
2\bigl(\Delta F_{\max}-\Delta F_{\min}\bigr).
\]
\end{corollary}

\begin{proof}
For $a<b$, note that for the threshold functions $h_a(z)=\mathbf{1}\{z\le a\}$ and $h_b(z)=\mathbf{1}\{z\le b\}$, their symmetric difference satisfies
\[
h_a(z)\oplus h_b(z)=\mathbf{1}\{a<z\le b\}.
\]
Conversely, every interval indicator $\mathbf{1}\{a<z\le b\}$ can be written as $h_a\oplus h_b$ for some $h_a,h_b\in H_{\mathrm{thr}}$. Therefore, by definition of the $H\Delta H$-divergence with base class $H_{\mathrm{thr}}$,
\[
d_{H_{\mathrm{thr}}\Delta H_{\mathrm{thr}}}(P,Q)
=
2\sup_{a<b}
\big|
P(a<Z\le b) - Q(a<Z\le b)
\big|.
\]

Using the identity $\{a<Z\le b\} = \{Z\le b\}\cap\{Z\le a\}^{c}$, we obtain
\[
P(a<Z\le b)=F_0(b)-F_0(a),
\qquad
Q(a<Z\le b)=F_1(b)-F_1(a),
\]
and hence
\[
d_{H_{\mathrm{thr}}\Delta H_{\mathrm{thr}}}(P,Q)
=
2\sup_{a<b}
\big|\Delta F(b)-\Delta F(a)\big|.
\]

Since $\Delta F_{\min} \le \Delta F(t) \le \Delta F_{\max}$ for all $t\in\mathbb{R}$,
we have
\[
\big|\Delta F(b)-\Delta F(a)\big|
\le
\Delta F_{\max}-\Delta F_{\min},
\qquad \forall\, a<b.
\]

Conversely, choose sequences $\{t_n^{\max}\}$ and $\{t_n^{\min}\}$ such that
$\Delta F(t_n^{\max})\to \Delta F_{\max}$ and
$\Delta F(t_n^{\min})\to \Delta F_{\min}$.
For each $n$, define
\[
a_n := \min\{t_n^{\max},t_n^{\min}\},
\qquad
b_n := \max\{t_n^{\max},t_n^{\min}\},
\]
so that $a_n<b_n$. Then
\[
\big|\Delta F(b_n)-\Delta F(a_n)\big|
=
\big|\Delta F(t_n^{\max})-\Delta F(t_n^{\min})\big|
\;\longrightarrow\;
\Delta F_{\max}-\Delta F_{\min}.
\]
Hence,
\[
\sup_{a<b}\big|\Delta F(b)-\Delta F(a)\big|
\ge
\Delta F_{\max}-\Delta F_{\min}.
\]

Combining the upper and lower bounds yields
\[
d_{H_{\mathrm{thr}}\Delta H_{\mathrm{thr}}}(P,Q)
=
2\bigl(\Delta F_{\max}-\Delta F_{\min}\bigr),
\]
which completes the proof.
\end{proof}

\begin{theorem}
\phantomsection
\label{appx:thm4}
Let $\mathcal{B}_0,\mathcal{B}_1$ be two finite sets of projected values and select $x\in\mathcal{B}_0$ and $y\in\mathcal{B}_1$ such that
\(
\delta_\mu(x-\tau)<0
\,\text{and}\,
\delta_\mu(y-\tau)>0
\)
for exchange between the two sets. Then any exchange rule does not decrease the $H\Delta H$-divergence for the threshold hypothesis class:
\[
d_{H_{\mathrm{thr}}\Delta H_{\mathrm{thr}}}^{\;\prime}(P',Q')
\;\;\ge\;\;
d_{H_{\mathrm{thr}}\Delta H_{\mathrm{thr}}}(P,Q),
\]
where $d_{H_{\mathrm{thr}}\Delta H_{\mathrm{thr}}}$ and $d_{H_{\mathrm{thr}}\Delta H_{\mathrm{thr}}}^{\;\prime}$ denote the divergence before and after the swap, respectively.
\end{theorem}

\begin{proof}
By Corollary~\ref{appx:cor1},
\[
d_{H_{\mathrm{thr}}\Delta H_{\mathrm{thr}}}(P,Q)
=
2\Big(\sup_{t\in\mathbb{R}}\Delta F(t)-\inf_{t\in\mathbb{R}}\Delta F(t)\Big),
\qquad
d_{H_{\mathrm{thr}}\Delta H_{\mathrm{thr}}}^{\;\prime}(P',Q')
=
2\Big(\sup_{t\in\mathbb{R}}\Delta F'(t)-\inf_{t\in\mathbb{R}}\Delta F'(t)\Big).
\]
Hence it suffices to show that the range of $\Delta F$ does not shrink:
\[
\Big(\sup \Delta F' - \inf \Delta F'\Big)\;\ge\;\Big(\sup \Delta F - \inf \Delta F\Big).
\]

By Lemma~\ref{appx:lemm1},
\[
F_0'(t)=F_0(t)-\delta_\mu\,\frac{1}{n_0}\,\mathbf{1}\{t\in I\},
\qquad
F_1'(t)=F_1(t)+\delta_\mu\,\frac{1}{n_1}\,\mathbf{1}\{t\in I\},
\qquad
\text{where}\quad I:=\bigl\{t\in\mathbb{R}:\ \delta_\mu x\le \delta_\mu t < \delta_\mu y\bigr\}.
\]
Therefore
\begin{equation}
\label{eq:deltaF-shift}
\Delta F'(t)
=
\Delta F(t)-\delta_\mu\,\kappa\,\mathbf{1}\{t\in I\},
\qquad \forall t\in\mathbb{R}.
\end{equation}
So the swap leaves $\Delta F$ unchanged outside $I$, and shifts it by a constant $-\delta_\mu\kappa$ on $I$.

Let $M:=\sup_{t}\Delta F(t)$ and $m:=\inf_{t}\Delta F(t)$, and similarly $M':=\sup_t \Delta F'(t)$ and $m':=\inf_t \Delta F'(t)$. Write the domain decomposition $\mathbb{R}=I\cup I^c$. From \eqref{eq:deltaF-shift},
\[
\sup_{t\in I^c}\Delta F'(t)=\sup_{t\in I^c}\Delta F(t),
\qquad
\inf_{t\in I^c}\Delta F'(t)=\inf_{t\in I^c}\Delta F(t),
\]
and
\[
\sup_{t\in I}\Delta F'(t)=\sup_{t\in I}\bigl(\Delta F(t)-\delta_\mu\kappa\bigr)
=\sup_{t\in I}\Delta F(t)-\delta_\mu\kappa,
\]
\[
\inf_{t\in I}\Delta F'(t)=\inf_{t\in I}\bigl(\Delta F(t)-\delta_\mu\kappa\bigr)
=\inf_{t\in I}\Delta F(t)-\delta_\mu\kappa.
\]
Hence
\[
M'=\max\Big\{\sup_{t\in I^c}\Delta F(t),\ \sup_{t\in I}\Delta F(t)-\delta_\mu\kappa\Big\},
\]
\[
m'=\min\Big\{\inf_{t\in I^c}\Delta F(t),\ \inf_{t\in I}\Delta F(t)-\delta_\mu\kappa\Big\}.
\]

Now consider two cases.

\smallskip
\noindent\emph{Case A: $\delta_\mu=+1$.}
Then $\Delta F$ is shifted \emph{downward} by $\kappa$ on $I$. Therefore $M'\ge \sup_{t\in I^c}\Delta F(t)$ and $m'\le \inf_{t\in I}\Delta F(t)-\kappa \le m-\kappa$ whenever the global infimum is attained on $I$; otherwise $m'=m$. In either situation, the maximum cannot increase, but the minimum cannot increase either; indeed the only possible change is that the function becomes \emph{no larger} on $I$. Consequently,
\[
M'-m'
\;\ge\;
M-m,
\]
because either the range is unchanged (if both extrema lie in $I^c$, or both lie in $I$), or the minimum decreases (if the infimum is attained on $I$), which enlarges the range.

\smallskip
\noindent\emph{Case B: $\delta_\mu=-1$.}
Then $\Delta F$ is shifted \emph{upward} by $\kappa$ on $I$. A symmetric argument shows that $m'$ cannot decrease while $M'$ cannot decrease; the only possible change is that the function becomes \emph{no smaller} on $I$. Hence again
\[
M'-m'
\;\ge\;
M-m.
\]

Combining the range inequality with Corollary~\ref{cor:cdf-hdh} yields
\[
d_{H_{\mathrm{thr}}\Delta H_{\mathrm{thr}}}^{\;\prime}(P',Q')
=
2(M'-m')
\;\ge\;
2(M-m)
=
d_{H_{\mathrm{thr}}\Delta H_{\mathrm{thr}}}(P,Q).
\]
This proves the claim.
\end{proof}

\newpage
\section{Analysis of the Gaussian Assumption}
\label{appx:gauss}
The Gaussian assumption in our analysis is used to obtain a tractable closed-form characterization of the quadratic term in the likelihood-ratio expression. To examine whether the key conclusion depends critically on this specific distributional choice, we consider an alternative Gamma-family model for the nonnegative behavior-distance variable $d$. We show that the stationary point induced by the Gamma likelihood ratio only introduces a higher-order perturbation relative to the Gaussian case. More importantly, the Gamma distribution preserves the conclusion in Thm.~\ref{thm:dstar-outside}.

\paragraph{Stationary point of the Gamma density ratio.}
Consider two Gamma distributions parameterized by shape $\alpha$ and rate $\beta$
\[
    f_1(d;\alpha_1,\beta_1) =\frac{\beta_1^{\alpha_1}}{\Gamma(\alpha_1)}d^{\alpha_1-1} e^{-\beta_1 d},
    \qquad
    f_2(d;\alpha_2,\beta_2) =\frac{\beta_2^{\alpha_2}}{\Gamma(\alpha_2)}d^{\alpha_2-1} e^{-\beta_2 d},
    \qquad 
    d>0.
\]
Their likelihood ratio can be formatted as
\[
 \frac{f_1(d)}{f_2(d)} =
    \frac{\Gamma(\alpha_2)\beta_1^{\alpha_1}}
    {\Gamma(\alpha_1)\beta_2^{\alpha_2}}
    d^{\alpha_1-\alpha_2}
    e^{-(\beta_1-\beta_2)d} =
    C_{\Gamma}
    d^{\alpha_1-\alpha_2}
    e^{-(\beta_1-\beta_2)d},
\]
where
\(
C_{\Gamma}=\frac{\Gamma(\alpha_2)\beta_1^{\alpha_1}}{\Gamma(\alpha_1)\beta_2^{\alpha_2}}
\)
is independent of $d$.

Taking the logarithm and differentiating with respect to \(d\) yields the stationary point
\[
d^\star_{\mathrm{Gamma}}=\frac{\alpha_1-\alpha_2}{\beta_1-\beta_2}=\frac{\mu_1^2\sigma_2^2-\mu_2^2\sigma_1^2}{\mu_1\sigma_2^2-\mu_2\sigma_1^2}, \, \text{by~}\alpha=\frac{\mu^2}{\sigma^2} \text{~and~} \beta=\frac{\mu}{\sigma^2}.
\]

\paragraph{Limited perturbation from Gamma modeling.}
Compared to its Gaussian counterpart in Eq.~\ref{eq:d}, the difference is
\[
\delta_p = d^\star_{\mathrm{Gamma}} - d^\star_{\mathrm{Gaussian}}=-\,\frac{(\mu_1-\mu_2)^2 \sigma_1^2 \sigma_2^2}{\left(\mu_1 \sigma_2^2-\mu_2 \sigma_1^2\right)\left(\sigma_2^2-\sigma_1^2\right)}.
\]
Under the non-degenerate condition that the denominator is bounded away from zero, this discrepancy is quadratic in the mean separation $(\mu_1-\mu_2)$. In Tab.~\ref{tab:gamma-gaussian}, we further show that adopting a Gamma-family model only introduces a limited perturbation, which supports the use of the Gaussian model as a tractable approximation in our framework.
\begin{table}[H]
\caption{The reported values are averaged over the last 50 checkpoints, recorded every 1000 training steps.}
\centering
\begin{tabular}{lccc}
\toprule
Dataset & $d^\star_{\mathrm{Gamma}}$ & $d^\star_{\mathrm{Gaussian}}$ & $|\delta_p|$ \\
\midrule
H-ME & 0.2849  & 0.3315 & 0.0465 \\
H-MR & 0.2028 & 0.2427 & 0.0399 \\
H-M  & 0.2899 & 0.3203 & 0.0303 \\
W2D-ME & 0.1537 & 0.1810 & 0.0273 \\
W2D-MR & 0.5786 & 0.6243 & 0.0457 \\
W2D-M & 0.3583 & 0.3919 & 0.0336 \\
\bottomrule
\end{tabular}
\label{tab:gamma-gaussian}
\end{table}

\paragraph{Preserved outside-interval property under Gamma modeling.}
Under the Gamma-family model, we have
\[
d^\star_{\mathrm{Gamma}} - \mu_1=\mu_2 \sigma_1^2 \frac{\mu_1-\mu_2} {\mu_1 \sigma_2^2-\mu_2 \sigma_1^2},\qquad d^\star_{\mathrm{Gamma}} - \mu_2=\mu_1 \sigma_2^2 \frac{\mu_1-\mu_2}{\mu_1 \sigma_2^2-\mu_2 \sigma_1^2}.
\]
These two quantities share the same sign, determined by
\(
\frac{\mu_1-\mu_2}{\mu_1 \sigma_2^2-\mu_2 \sigma_1^2}.
\)
Hence,
\[
d^\star_{\mathrm{Gamma}} \notin 
\big(\min\{\mu_1,\mu_2\},\max\{\mu_1,\mu_2\}\big).
\]
Therefore, the conclusion of Thm.~\ref{thm:dstar-outside} remains unchanged under the Gamma-family model.

\newpage
\section{DARE Instantiation Details}
\label{appx:insta}
We integrate DARE into two representative offline RL algorithms, Cal-QL~\citep{nakamoto2023cal} and IQL\citep{kostrikov2022offline}, by applying differentiated updates to the exchanged batches \({b'}_{off}\) and \({b'}_{on}\). For offline-like samples in \({b'}_{off}\), we still use the conservative objectives to preserve stability. For online-like samples in \({b'}_{on}\), we relax these constraints to facilitate adaptation.

\paragraph{DARE-Cal-QL.} We extend Cal-QL with DARE, resulting in the variant DARE-Cal-QL (DARE-C). Cal-QL applies a conservative regularizer \(\mathcal{R}\) uniformly to all samples to mitigate value overestimation. In DARE-C, this penalty is applied selectively: \(\mathcal{R}\) is retained for offline-like samples in \({b'}_{off}\) to preserve stability, but removed for online-like samples in \({b'}_{on}\) to facilitate adaptation. The original objective becomes
\begin{equation}
\begin{aligned}
\mathcal{L}_{\text{DARE}}^{\text{Cal-QL}}(Q)
= \mathbb{E}_{(s,a,s')\sim\mathcal{D}}&\Big[
\mathds{1}_{\{(s,a,s')\in {b'}_{off}\}}\big(
(Q(s,a)- \hat{\mathcal{B}}^\pi \hat{Q}_{\text{target}}(s,a))^2+\alpha\mathcal{R}
\big)\\
&+\ \mathds{1}_{\{(s,a,s')\in {b'}_{on}\}}
\big(Q(s,a)-\hat{\mathcal{B}}^\pi \hat{Q}_{\text{target}}(s,a)\big)^2
\Big].
\end{aligned}
\label{eq:calql-strat}
\end{equation}

\paragraph{DARE-IQL.}
\label{c:iql}
When applied to IQL, DARE brings in a variant DARE-IQL (DARE-I). IQL trains the Q-function via value regression and updates the policy through advantage-weighted regression. In DARE-I, these updates are differentiated: for offline-like samples \({b'}_{off}\), we retain the original value targets and use the PEX objective for policy learning; for online-like samples \({b'}_{on}\), we replace the value targets with TD-based maximum-Q estimates and switch the policy update to an entropy-regularized SAC objective~\citep{haarnoja2018soft}. The original Q-function and policy updates become:
\begin{equation}
\begin{aligned}
\mathcal{L}_{\text{DARE}}^{\text{IQL}}(Q) = \mathbb{E}_{(s,a,s') \sim \mathbf{D}}& \Big[
\mathds{1}_{\{(s,a,s') \in {b'}_{\text{off}}\}} \big(r(s,a) + \gamma V(s') - Q(s,a) \big)^2 \\
&+\
\mathds{1}_{\{(s,a,s') \in {b'}_{\text{on}}\}} \big(r(s,a) + \gamma \max_{a'} \hat{Q}_{\text{target}}(s', a') - Q(s,a) \big)^2 \Big].
\end{aligned}
\label{eq:iql-q-strat}
\end{equation}

For policy optimization, the overall policy objective is given by
\begin{equation}
\begin{aligned}
\mathcal{L}_{\text{DARE}}^{\text{IQL}}(\pi)
=\ & \mathbb{E}_{(s,a)\sim\mathbf{D},\tilde a \sim \tilde \pi(s)}\Big[
\mathds{1}_{\{(s,a)\in {b'}_{\text{off}}\}}
w(s, \tilde a)\log \pi(\tilde a\mid s)\Big]\\
&+\,\mathbb{E}_{s\sim\mathbf{D},\tilde a\sim \pi(\cdot\mid s)}\Big[
\mathds{1}_{\{(s,\tilde a)\in {b'}_{\text{on}}\}}
\big(\alpha\log \pi(\tilde a\mid s)-Q(s,\tilde a)\big)\Big],
\end{aligned}
\label{eq:iql-policy-strat}
\end{equation}
where $\tilde \pi$ is the composite policy defined in PEX with $w(s,\tilde a) = \exp(\beta(Q(s,\tilde a)-V(s)))$, and entropy coefficient $\alpha$ is set to $0.2$ for locomotion tasks and $0.02$ for AntMaze tasks.

\paragraph{Energy-Guided Diffusion Model as Behavior Policy.}
Energy-guided diffusion model has been widely deployed to improve data quality or sample efficiency~\citep{lu2023contrastive,liu2024energy,zhang2024object,xu2025energy}. Mathematically, we take the pretrained diffusion behavior model $\mu_0$ as the base distribution and use the Q-function as the energy term, which defines an induced policy $\pi_0$. We have the marginal distributions $\mu_t$ and $\pi_t$ of the noise-perturbed action $a_t$ satisfying:
\begin{equation}
    \pi_t(\mathbf{a}_t \mid \mathbf{s}) \propto \mu_t(\mathbf{a}_t \mid \mathbf{s}) \, e^{\mathcal{E}_t(\mathbf{s}, \mathbf{a}_t)}.
    \label{eq:diffusion}
\end{equation}
$\mathcal{E}_t(s, a_t)$ is an intermediate energy function determined by the learned action evaluation model $Q(s, a_0)$. Specifically, $\mathcal{E}_t(s, a_t) = \log \mathbb{E}_{\mu_{t}(a_0 \mid a_t, s)} \left[ e^{\beta Q(s, a_0)} \right]$, and $\mathcal{E}_0(s, a_0) = \beta Q(s, a_0)$. To allow action sampling from $\pi_0$, the score function can be expressed as:
\begin{equation}
\nabla_{a_t} \log \pi_t(a_t \mid s) = 
\underbrace{\nabla_{a_t} \log \mu_t(a_t \mid s)}_{\approx -\epsilon_\theta(a_t \mid s, t)/\sigma_t} 
+ \underbrace{\nabla_{a_t} \mathcal{E}_t(s, a_t)}_{\approx f_\phi(s, a_t, t)},
\label{eq:sampling}
\end{equation}
where $\epsilon_\theta(a_t \mid s, t)/\sigma_t$ denotes the state-conditinal diffusion model to model the behavior policy $\mu_0$. The energy guidance model $f_{\phi}$ employs a contrastive learning objective based on self-normalized energy labels that defined as:
\begin{equation}
\label{eq:cep-loss}
\min_{\phi} \ \mathbb{E}_{p(t)} \, \mathbb{E}_{q_0(x_0^{(1:K)})} \, \mathbb{E}_{p(\epsilon^{(1:K)})} \left[
- \sum_{i=1}^{K}
\frac{e^{-\beta \mathcal{E}(x_0^{(i)})}}{\sum_{j=1}^K e^{-\beta \mathcal{E}(x_0^{(j)})}}
\log \frac{e^{-f_\phi(x_t^{(i)}, t)}}{\sum_{j=1}^K e^{-f_\phi(x_t^{(j)}, t)}}
\right].
\end{equation}
Here, each perturbed sample is generated by the forward diffusion process: \( x_t^{(i)} = \alpha_t x_0^{(i)} + \sigma_t \epsilon^{(i)} \), with constants \( \alpha_t, \sigma_t \) defined by the diffusion schedule.

\newpage
\section{Experimental Details}
\label{appx:impl}
In our experiments, we evaluate DARE on two standard D4RL benchmarks~\citep{fu2020d4rl}: MuJoCo Locomotion and AntMaze Navigation. For instantiation, we extend Cal-QL~\citep{nakamoto2023cal} to DARE-C and IQL~\citep{kostrikov2022offline} to DARE-I. In the CQL-based group, we compare against CQL~\citep{kumar2020conservative}, Cal-QL, and EDIS-C~\citep{liu2024energy}, a Cal-QL variant. In the IQL-based group, we include IQL, PEX~\citep{zhang2023policy}, and EDIS-I, the IQL counterpart of EDIS-C. All models are first trained offline for 1M steps and then fine-tuned online for 0.2M steps. The results are averaged over the last four evaluations and five random seeds. For a fair comparison, \textbf{all methods are initialized from the same offline-trained models, using Cal-QL models for EDIS-C and DARE-C and IQL models for PEX, EDIS-I, and DARE-I.}

\subsection{Hyperparameters for DARE}
DARE introduces only one additional hyperparameter: the update frequency for the class-conditional mean $\mu_c$ and variance $\sigma_c$ of the deviation statistic $d(s,a)$. We update $(\mu_c,\sigma_c)$ every 1000 training steps by sampling 10{,}000 samples from the offline dataset and 10{,}000 from the online replay buffer, computing $d(s,a)$ via the reference behavior model, and estimating the statistics separately for each source.

\subsection{Hyperparameters for CQL and Cal-QL}
We implement the CQL and Cal-QL based on \url{https://github.com/tinkoff-ai/CORL}, and primarily follow the authors' recommended hyperparameters~\citep{tarasov2023corl}. Since CQL-based algorithms are known to be sensitive to hyperparameter choices, we provide the exact settings in our experiments to facilitate the reproducibility. Please refer to Tab.~\ref{tab:cql-hyperparameters} for the details about the hyperparameters in our CQL-based implementation.

\begin{table}[h]
\centering
\caption{Hyperparameters in CQL-based implementation.}
\begin{tabular}{lll}
\toprule
\textbf{Hyperparameter} & \textbf{Mojoco locomotion} & \textbf{AntMaze navigation}\\
\midrule
\multicolumn{2}{l}{\textbf{General Settings}} & \\
Replay buffer size & 2,000,000 & 2,000,000 \\
Batch size & 256 & 256 \\
Discount factor \(\gamma\) & 0.99 & 0.99 \\
Reward scale / bias & 1.0 / 0.0 & 10.0 / -5.0 \\
Normalize states & True & False \\
Normalize reward & False & True \\
Orthogonal initialization & True & True\\
Is sparse reward & False & True \\
\midrule
\multicolumn{2}{l}{\textbf{CQL Hyperparameters}} & \\
Policy learning rate & \(1 \times 10^{-4}\) & \(1 \times 10^{-4}\) \\
Critic learning rate & \(3 \times 10^{-4}\) & \(3 \times 10^{-4}\) \\
Soft target update rate \(\tau\) & 0.005 & 0.005 \\
Target update period & 1 & 1 \\
Automatic entropy tuning & True & True \\
Backup entropy & False & False \\
CQL regularization (\(\alpha\), offline / online) & 10.0 / 10.0 & 5.0 / 5.0 \\
CQL Lagrange & False & True\\
CQL temperature & 1.0 & 1.0 \\
Target action gap & \(-1.0\) & 0.8 \\
Max target backup & False & True \\
Clip diff range & \([-200,\ \infty)\) & \([-200,\ \infty)\) \\
Importance sampling & True & True \\
\midrule
\multicolumn{2}{l}{\textbf{Network Architecture}} & \\
Q-network hidden layers & 3 & 5 \\
Hidden dimension (actor / critic) & 256 / 256 & 256 / 256 \\
\bottomrule
\end{tabular}
\label{tab:cql-hyperparameters}
\end{table}

\newpage
\subsection{Hyperparameters for IQL and PEX}
We implement IQL and PEX based on \url{https://github.com/Haichao-Zhang/PEX}, the hyperparameters of which is illustrated in Tab.~\ref{tab:iql-hyperparams}.
\begin{table}[h]
\centering
\caption{Hyperparameters for the IQL-based experiments.}
\begin{tabular}{lc}
\toprule
\textbf{Hyperparameter} & \textbf{Value} \\
\midrule
Discount factor \(\gamma\) & 0.99 \\
Hidden dimension & 256 \\
Number of hidden layers & 2 \\
Batch size & 256 \\
Learning rate & \(3 \times 10^{-4}\) \\
Target update rate & 0.005 \\
Expectile parameter \(\tau\) & 0.9, AntMaze / 0.7, Locomotion \\
Inverse temperature \(\beta\) & 10.0, AntMaze / 3.0, Locomotion  \\
\bottomrule
\end{tabular}
\label{tab:iql-hyperparams}
\end{table}

\subsection{Hyperparameters for EDIS}
The implementation of EDIS are referred to \url{https://github.com/liuxhym/EDIS}. For EDIS-C, we use its official implementation of Cal-QL. For EDIS-I, we modify the classes of Q-function, value function, and policy function to match those in IQL and PEX, so that the same offline models can be loaded for initialization. The hyperparameters used in the EDIS module remain unchanged and, for convenience, are detailed in Tab.~\ref{tab:edis-hyperparams}.
\begin{table}[h]
\centering
\caption{Hyperparameters in EDIS.}
\label{tab:edis-hyperparams}
\begin{tabular}{lc}
\toprule
\textbf{Hyperparameter} & \textbf{Value} \\
\midrule
Network Type (Denoising) & Residual MLP \\
Denoising Network Depth & 6 layers \\
Denoising Steps & 128 steps \\
Denoising Network Learning Rate & $3 \times 10^{-4}$ \\
Denoising Network Hidden Dimension & 1024 units \\
Denoising Network Batch Size & 256 samples \\
Denoising Network Activation Function & ReLU \\
Denoising Network Optimizer & Adam \\
Learning Rate Schedule (Denoising Network) & Cosine Annealing \\
Training Epochs (Denoising Network) & 50,000 epochs \\
Training Interval Environment Step (Denoising Network) & Every 10,000 steps \\
Energy Network Hidden Dimension & 256 units \\
Negative Samples (Energy Network Training) & 10 \\
Energy Network Learning Rate & $1 \times 10^{-3}$ \\
Energy Network Activation Function & ReLU \\
Energy Network Optimizer & Adam \\
\bottomrule
\end{tabular}
\end{table}

The guidance scale \(s\) follows the empirical settings from prior work~\citep{lu2023contrastive}, as summarized in Tab.~\ref{tab:guidance-scale}.

\begin{table}[H]
\centering
\caption{Guidance scale \(s\) across different environments.}
\begin{tabular}{lc}
\toprule
\textbf{Dataset} & \textbf{Guidance Scale \(s\)} \\
\midrule
walker2d-medium-expert-v2 & 5.0 \\
halfcheetah-medium-expert-v2 & 3.0 \\
hopper-medium-expert-v2 & 2.0 \\
walker2d-medium-replay-v2 & 5.0 \\
halfcheetah-medium-replay-v2 & 8.0 \\
hopper-medium-replay-v2 & 3.0 \\
walker2d-medium-v2 & 10.0 \\
halfcheetah-medium-v2 & 10.0 \\
hopper-medium-v2 & 8.0 \\
antmaze-umaze-v2 & 3.0 \\
antmaze-medium-play-v2 & 4.0 \\
antmaze-large-play-v2 & 2.0 \\
antmaze-umaze-diverse-v2 & 1.0 \\
antmaze-medium-diverse-v2 & 3.0 \\
antmaze-large-diverse-v2 & 3.0 \\
\bottomrule
\end{tabular}
\label{tab:guidance-scale}
\end{table}

\subsection{Offline Model Performance}
\label{sec_apd:offline_initial}
Before fine-tuning, we present the performance of the offline-trained models on both Locomotion and AntMaze benchmarks. These results serve as the initialization for all compared methods, ensuring that performance differences in the online phase come solely from the fine-tuning strategy rather than the quality of the offline implementation codebases. Tab.~\ref{tab:offline-performance} summarizes the scores of CQL, Cal-QL and IQL across different datasets.
\begin{table}[H]
\centering
\caption{Offline training performance before online fine-tuning.}
\label{tab:offline-performance}
\begin{tabular}{l|ccc}
\toprule
\textbf{Dataset} & \textbf{CQL} & \textbf{Cal-QL} & \textbf{IQL} \\
\midrule
HC-ME   & 90.4\footnotesize{\( \pm \)3.5} & 82.9\footnotesize{\( \pm \)4.1} & 90.2\footnotesize{\( \pm \)1.4} \\
H-ME    & 108.1\footnotesize{\( \pm \)2.4} & 109.0\footnotesize{\( \pm \)2.1} & 32.8\footnotesize{\( \pm \)32.6} \\
W2D-ME  & 109.7\footnotesize{\( \pm \)0.2} & 108.6\footnotesize{\( \pm \)0.8} & 108.3\footnotesize{\( \pm \)2.0} \\
HC-MR   & 45.5\footnotesize{\( \pm \)0.2} & 46.5\footnotesize{\( \pm \)0.2} & 43.9\footnotesize{\( \pm \)0.4} \\
H-MR    & 90.2\footnotesize{\( \pm \)10.5} & 64.8\footnotesize{\( \pm \)27.1} & 84.2\footnotesize{\( \pm \)13.6} \\
W2D-MR  & 78.6\footnotesize{\( \pm \)4.5} & 85.0\footnotesize{\( \pm \)2.9} & 70.8\footnotesize{\( \pm \)4.1} \\
HC-M    & 47.1\footnotesize{\( \pm \)0.3} & 48.3\footnotesize{\( \pm \)0.3} & 48.1\footnotesize{\( \pm \)0.1} \\
H-M     & 63.9\footnotesize{\( \pm \)1.5} & 61.9\footnotesize{\( \pm \)2.0} & 56.2\footnotesize{\( \pm \)4.4} \\
W2D-M   & 81.9\footnotesize{\( \pm \)1.2} & 83.7\footnotesize{\( \pm \)1.2} & 74.0\footnotesize{\( \pm \)3.1} \\
\midrule
total (L) & 715.4 & 690.7 & 608.5 \\
\midrule
AM-LD   & 30.6\footnotesize{\( \pm \)5.6} & 34.8\footnotesize{\( \pm \)13.4} & 35.2\footnotesize{\( \pm \)10.0} \\
AM-LP   & 32.7\footnotesize{\( \pm \)4.1} & 42.4\footnotesize{\( \pm \)9.7} & 36.6\footnotesize{\( \pm \)3.2} \\
AM-MD   & 57.6\footnotesize{\( \pm \)3.5} & 66.2\footnotesize{\( \pm \)6.1} & 70.4\footnotesize{\( \pm \)5.2} \\
AM-MP   & 64.8\footnotesize{\( \pm \)6.1} & 68.4\footnotesize{\( \pm \)3.8} & 74.6\footnotesize{\( \pm \)4.7} \\
AM-UD   & 28.2\footnotesize{\( \pm \)23.9} & 46.0\footnotesize{\( \pm \)20.1} & 60.2\footnotesize{\( \pm \)6.4} \\
AM-U    & 87.6\footnotesize{\( \pm \)4.6} & 76.6\footnotesize{\( \pm \)2.7} & 91.4\footnotesize{\( \pm \)0.8} \\
\midrule
total (AM) & 301.5 & 334.4 & 368.4 \\
\midrule
\midrule
total & 1016.9 & 1025.1 & 976.9 \\
\bottomrule
\end{tabular}
\end{table}

\newpage
\section{Comparisons of Online Fine-tuning Processes}
\label{appx:main}
\begin{figure}[h]
    \centering
    \includegraphics[width=0.8\linewidth]{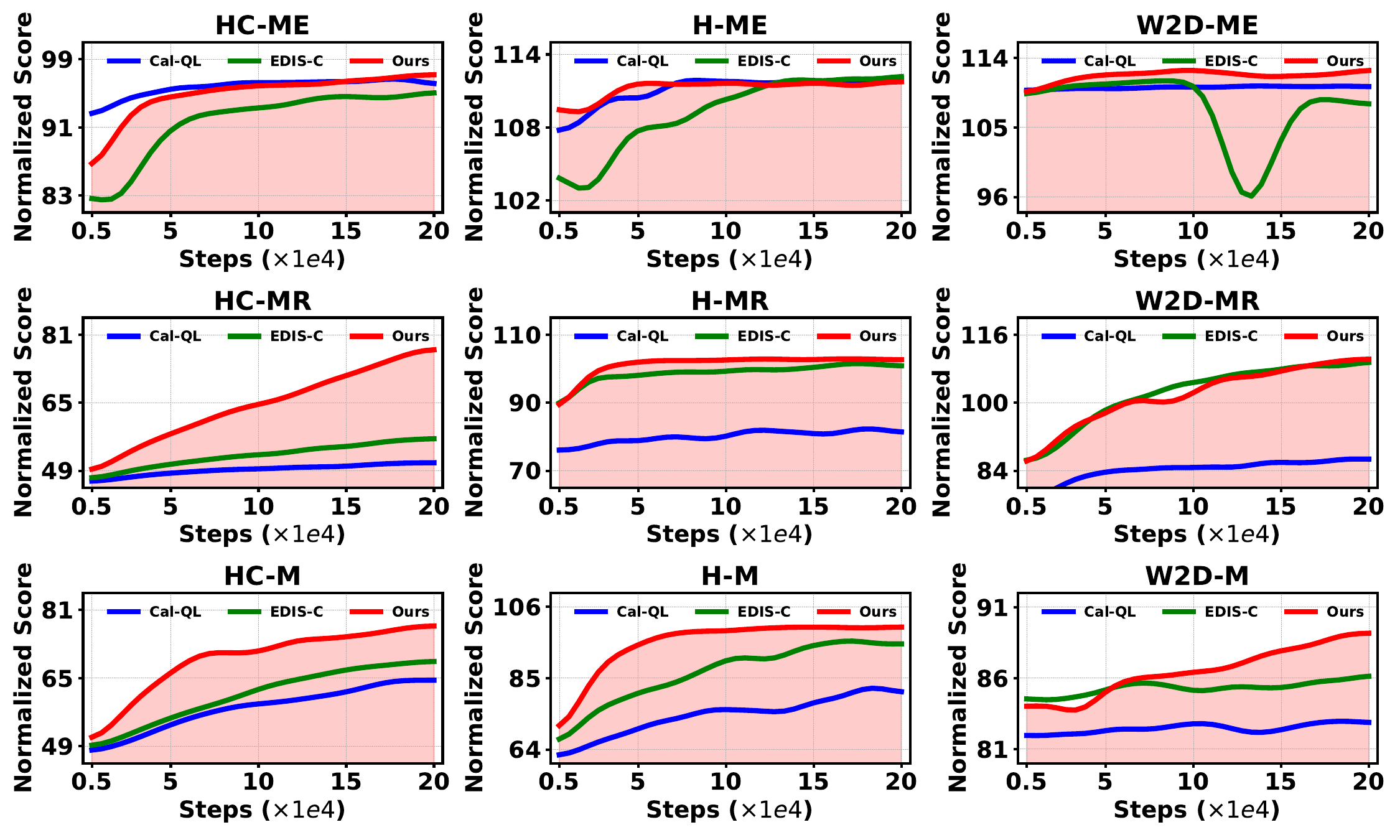}
    \caption{\small Online training curves on Locomotion benchmarks comparing Cal-QL, EDIS-C, and Ours.}
    \label{fig:online-cql}
\end{figure}
Complementary to the main results in the paper, we depict the online training dynamics across algorithms. As shown in Figs.~\ref{fig:online-cql}  our method consistently outperforms both Cal-QL and EDIS-C, while exhibiting smoother training trajectories.
\begin{figure}[H]
    \centering
    \includegraphics[width=0.8\linewidth]{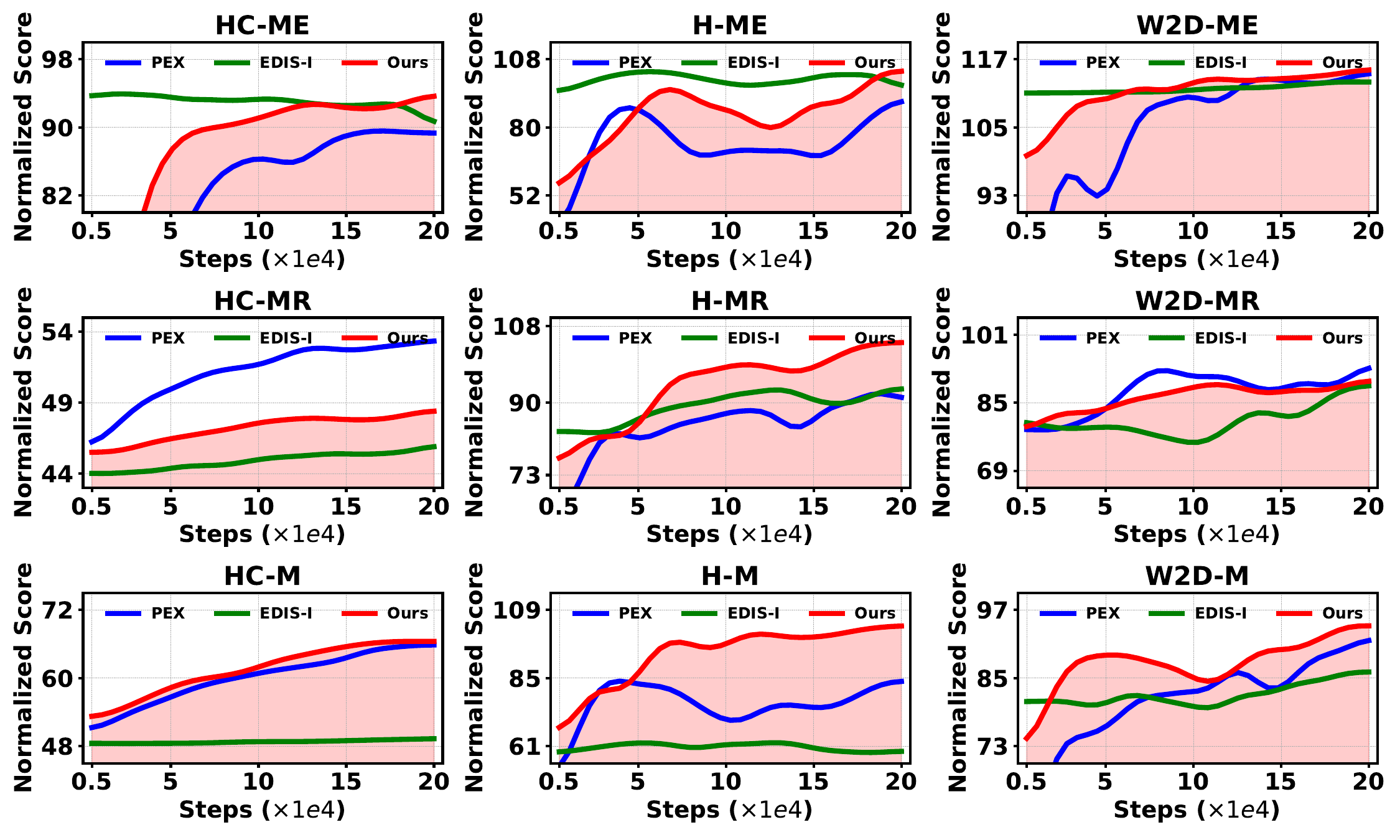}
    \caption{\small Online training curves on Locomotion benchmarks comparing PEX, EDIS-I, and Ours.}
    \label{fig:online-iql}
\end{figure}
As shown in Figs.~\ref{fig:online-iql}, our method consistently exhibits smoother optimization dynamics and higher asymptotic returns than both PEX and EDIS-I across most benchmarks, with the exceptions of \textit{halfcheetah-medium-replay} and \textit{walker2d-medium-replay}. The curves demonstrates that EDIS-I attains strong initial performance on expert datasets and effectively exploits high-quality offline data at early stages. However, its performance declines as the optimization process continues, which suggests limited robustness during the online adaptation phase. 

\newpage
\section{Additional Results for Main Table}
\label{appx:additional-results}
This section provides additional experimental results to complement the main results reported in Tab.~\ref{tab:performance}. 

\paragraph{Results on Kitchen tasks.}
We evaluate DARE on the Kitchen tasks~\citep{gupta2020relay} under the IQL-based setting. As shown in Tab.~\ref{tab:kitchen}, DARE-I consistently outperforms both IQL and PEX, which supports the effectiveness of DARE. 
\begin{table}[H]
\centering
\caption{Results on Kitchen tasks under the IQL-based setting with 0.5M offline training steps and 1M online fine-tuning steps.}
\label{tab:kitchen}
\begin{tabular}{lccc}
\toprule
Task & IQL & PEX & DARE-I \\
\midrule
\texttt{kitchen-partial-v0} & 42.29 & 44.17 & \textbf{50.21} \\
\texttt{kitchen-mixed-v0}   & 53.12 & 53.75 & \textbf{56.25} \\
\bottomrule
\end{tabular}
\end{table}

\paragraph{Results on different instantiations.}
We examine the generality of the exchange mechanism on the Locomotion tasks under different online-objective instantiations. In these experiments, the exchange mechanism is kept fixed, while the online objective applied to online-like samples is replaced with different alternatives. As shown in Tab.~\ref{tab:instan-loco}, enabling exchange consistently improves performance over the corresponding no-exchange variants across all instantiations, demonstrating that the proposed mechanism is not tied to a specific online objective.
\begin{table}[H]
\centering
\caption{Locomotion results for different instantiations of DARE-I. w/o S (no SAC), w/o P (no PEX), w/o S\&P (TD-only), with w/o E. and w/ E. indicating without and with exchange.}
\begin{tabular}{l|cc|cc|cc}
\toprule
M: medium
& \multicolumn{2}{c|}{DARE-I.w/o S}
& \multicolumn{2}{c|}{DARE-I.w/o P}
& \multicolumn{2}{c}{DARE-I.w/o S\&P} \\
MR: medium-replay
& w/o E. & w/ E.
& w/o E. & w/ E.
& w/o E. & w/ E. \\
\midrule
M(total)    & 231.1 & \textbf{238.4} & 207.2 & \textbf{230.8} & 232.4 & \textbf{247.9}\\
MR(total)   & 259.2 & \textbf{261.4} & 215.4 & \textbf{227.0} & 223.3 & \textbf{229.5}\\
\bottomrule
\end{tabular}
\label{tab:instan-loco}
\end{table}

\paragraph{Results on the TD3+BC backbone.}
The main results in Tab.~\ref{tab:performance} are primarily based on CQL and IQL backbones. To further examine the generality of DARE, we extend the proposed exchange mechanism to TD3+BC~\citep{fujimoto2021minimalist}, another representative offline RL algorithm that regularizes the TD3 actor update with a behavior cloning (BC) term
\[
    \pi
    =
    \arg\max_{\pi}
    \mathbb{E}_{(s,a)\sim \mathcal{D}}
    \left[
        \lambda Q(s,\pi(s))
        -
        \left\|\pi(s)-a\right\|_2^2
    \right],
\]
where $\lambda$ is a weighting coefficient that balances Q-value maximization and behavior cloning regularization.

For evaluation, we apply DARE to two TD3+BC-based O2O methods, BOORL~\citep{hu2024bayesian} and LAROO~\citep{guo2026tackling}. During online fine-tuning, the offline objective active by applying it once every two gradient updates, so that the actor remains regularized by identified offline data via DARE. In addition, although the official implementations of BOORL and LAROO use a UTD ratio of 5, we set the UTD ratio to 1 in our experiments to match the training setting used throughout this paper and ensure a fair comparison. As shown in Tab.~\ref{tab:td3bc}, DARE improves the corresponding baselines, further indicating that the proposed mechanism can generalize beyond CQL- and IQL-based settings.
\begin{table}[H]
\centering
\caption{Locomotion results for.}
\begin{tabular}{l|cc|cc}
\toprule
& \multicolumn{2}{c|}{BOORL}
& \multicolumn{2}{c}{LAROO} \\
& base & +DARE
& base & +DARE \\
\midrule
HC-M     & 91.9 & \textbf{93.9}   & 92.5 & \textbf{94.1} \\
HC-MR    & 82.6 & \textbf{84.1}   & \textbf{80.2} & 78.5 \\
H-M      & 108.1 & 108.2          & 109.0 & \textbf{110.0} \\
H-MR     & 104.2 & \textbf{109.8} & 107.9 & \textbf{108.6} \\
W2D-M    & 97.6 & \textbf{100.9}  & 106.9 & \textbf{111.9} \\
W2D-MR   & 103.1 & \textbf{108.1} & 105.6 & \textbf{109.1} \\
\bottomrule
\end{tabular}
\label{tab:td3bc}
\end{table}

\newpage
\section{Visualization of Exchange Behavior}
\label{appx:exchange}
\begin{figure}[H]
    \centering
    \includegraphics[width=0.92\linewidth]{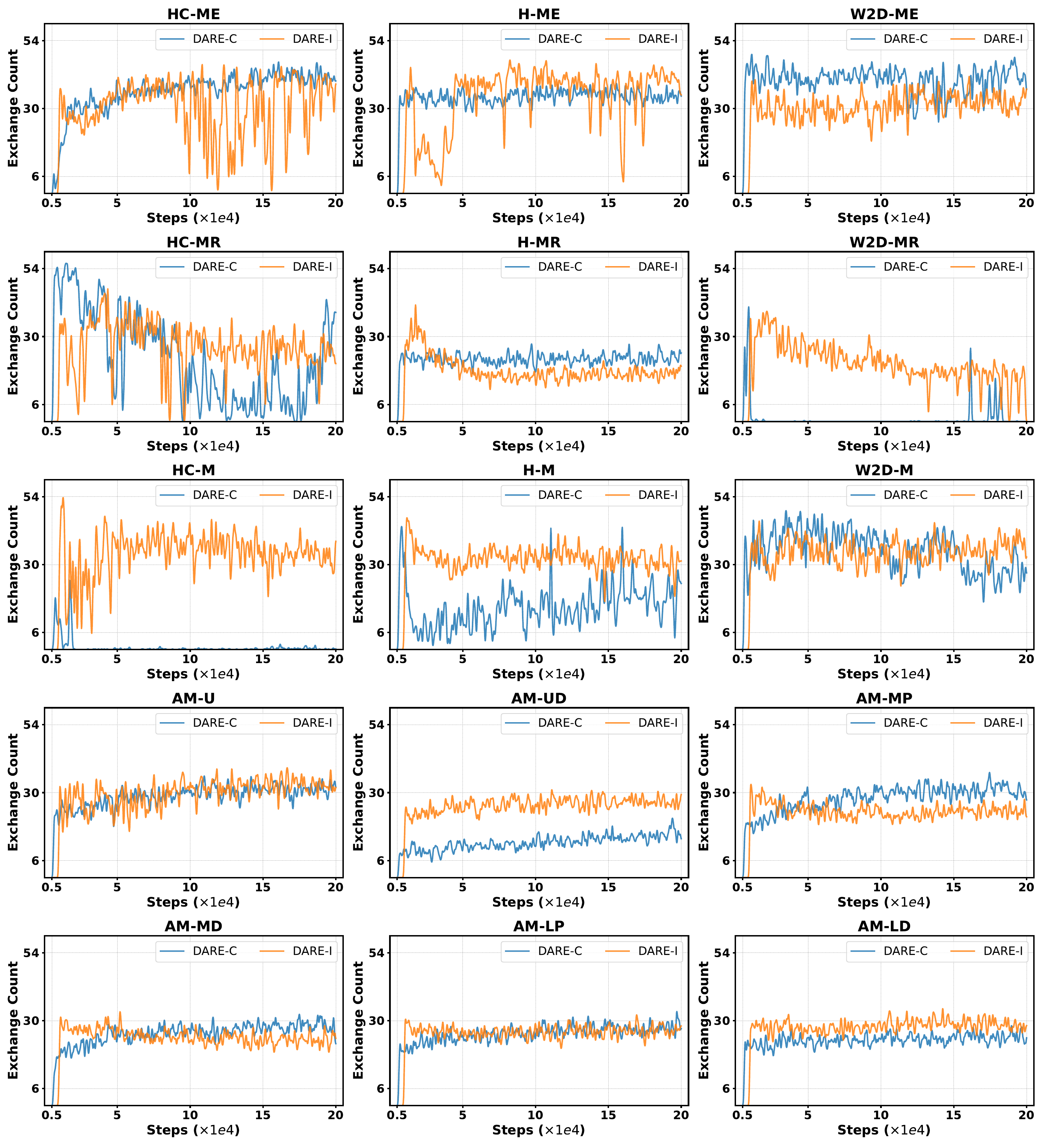}
    \caption{\small Adaptive exchange behavior of DARE-C and DARE-I during online training across different tasks.}
    \label{fig:exchange-apd}
\end{figure}

We present the exchange behavior of DARE-C and DARE-I during online training across all tested environments. We depict the number of exchanges over training iterations to characterize how often the proposed exchange mechanism triggers sample reassignment in practice, as shown in Fig.~\ref{fig:exchange-apd}.

\newpage
\section{Further Results and Analysis on Adroit}
\label{appx:adroit}
We study how different online update schemes interact with Adroit manipulation tasks under a fixed exchange mechanism. PEX constrains online interaction at the behavior level rather than through an explicit objective. Following \citet{zhang2023policy}, the behavior policy is constructed by composing the frozen offline policy $\pi_\beta$ and the online policy $\pi_\theta$, with actions selected according to a critic-induced distribution
\(
P(i\mid s)\propto \exp\!\bigl(Q(s,a_i)/\alpha\bigr),
a_i\sim\{\pi_\beta(s),\pi_\theta(s)\}.
\)
As a result, executed actions remain restricted to proposals from $\pi_\beta$ and $\pi_\theta$. When the offline initialization is weak, this restriction limits deviation from suboptimal contact behaviors in Adroit tasks.

Entropy-regularized SAC objectives exhibit different characteristics. Adroit tasks require fine-grained, contact-rich control, and high action entropy does not consistently align with these requirements~\citep{fu2020d4rl}. At the same time, SAC promotes deviation from offline actions and introduces exploratory behavior that is absent from PEX-based updates, which can be beneficial when offline coverage is poor.

\begin{figure}[H]
    \centering
    \includegraphics[width=0.85\linewidth]{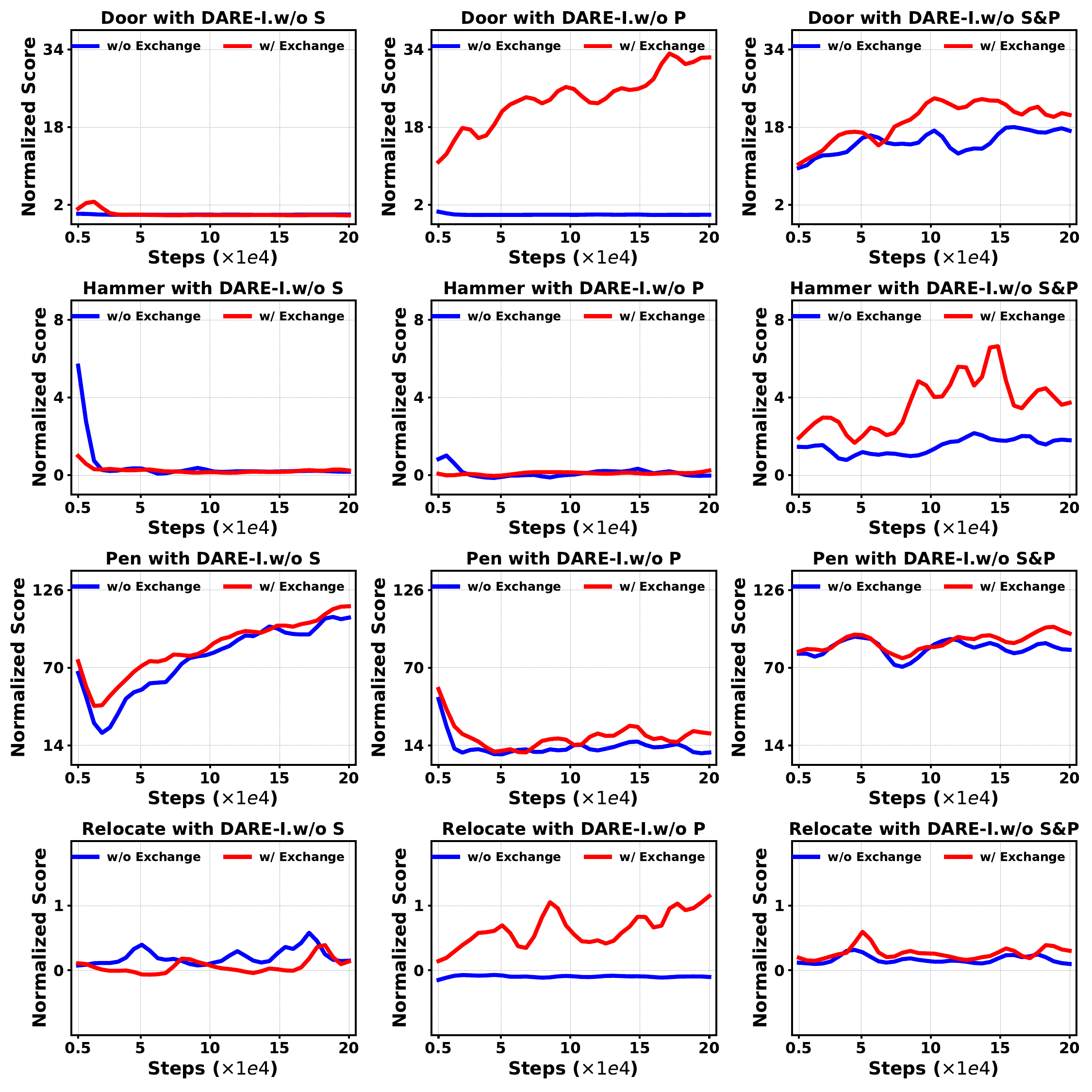}
    \caption{\small Training curves on Adroit manipulation tasks.}
    \label{fig:adroit}
\end{figure}

Taken together, these results indicate that the effectiveness of online objective instantiations depends on task characteristics and the quality of offline initialization. DARE therefore adopts a objective-agnostic exchange mechanism, which allows different online objectives to be instantiated without modifying the underlying sample reassignment rule. In Fig.~\ref{fig:adroit}, the exchange-based method exhibits stronger overall performance.
\newpage

\section{Alternative Behavior Model Instantiations}
\label{appx:behavior}
\begin{figure}[H]
    \centering
    \includegraphics[width=0.85\linewidth]{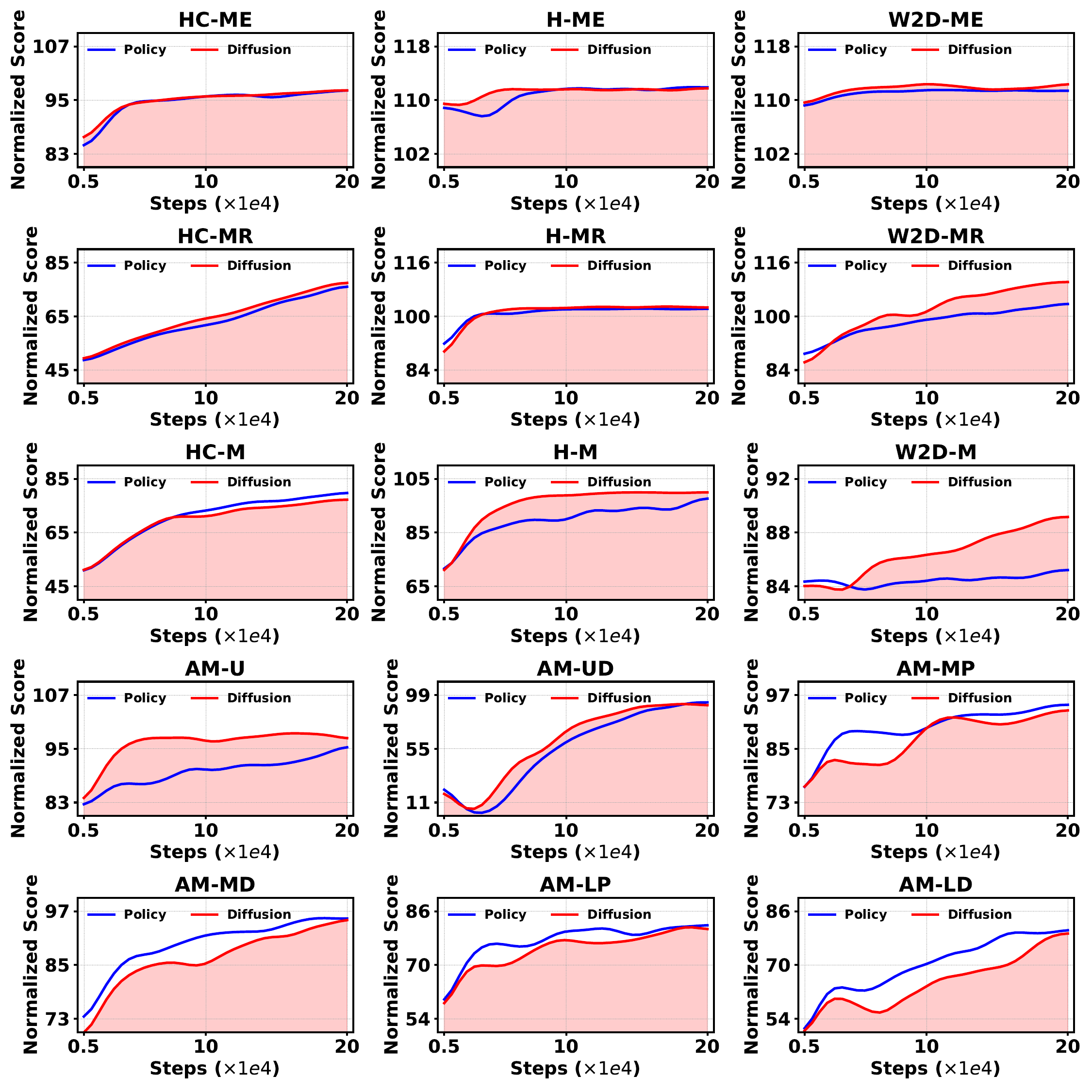}
    \caption{\small Fine-tuning curves with different behavior model instantiations for DARE-C.}
    \label{fig:cql_on}
\end{figure}

\begin{figure}[H]
    \centering
    \includegraphics[width=0.9\linewidth]{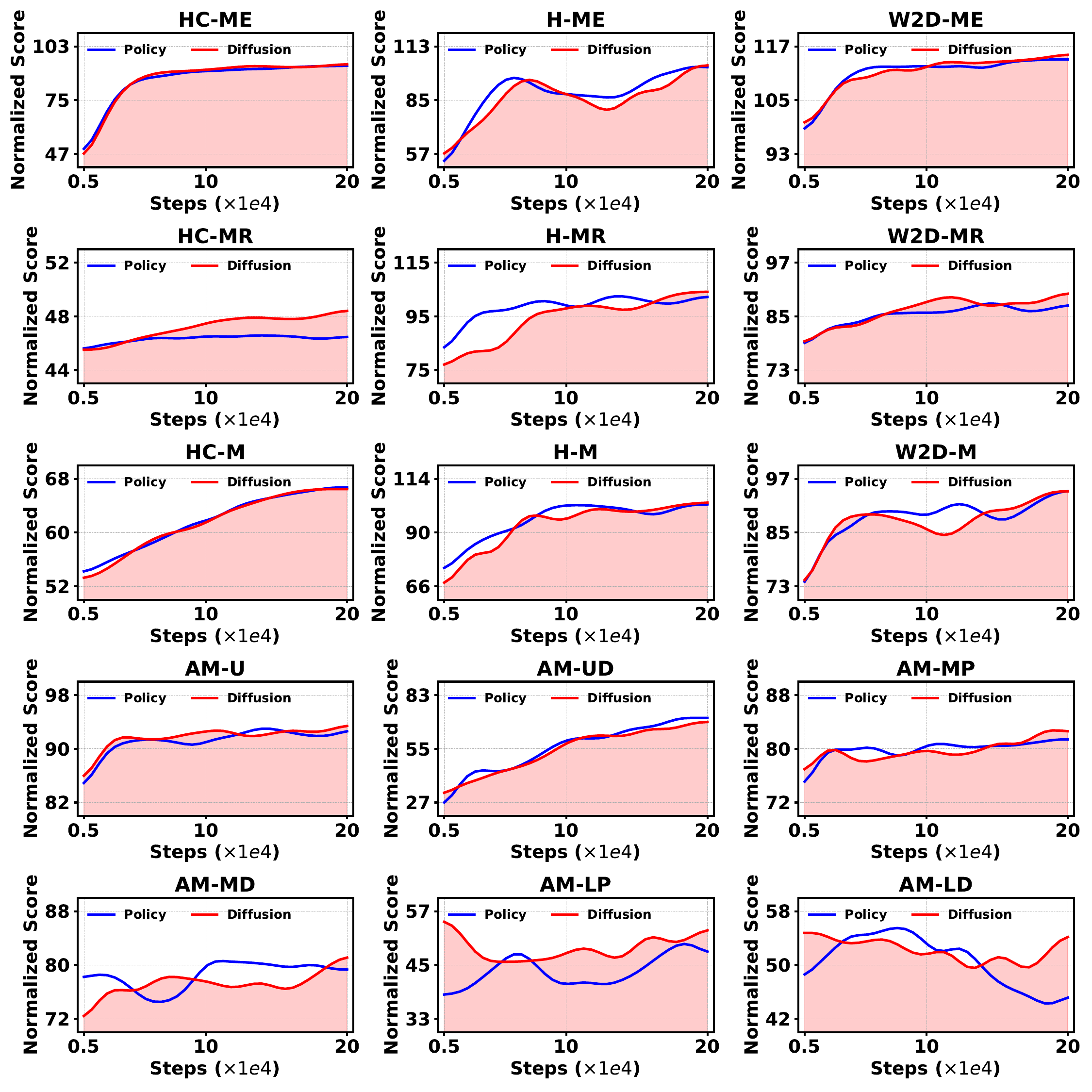}
    \caption{\small Fine-tuning curves with different behavior model instantiations for DARE-I.}
    \label{fig:iql_on}
\end{figure}

We compare two behavior model instantiations in DARE. We consider a diffusion-based behavior model trained on the offline dataset and an alternative instantiation that uses the online policy actor as a consistency model for online behaviors. Fig.~\ref{fig:cql_on} and Fig.~\ref{fig:iql_on} present the fine-tuning curves for DARE-I and DARE-C under these two behavior model choices. In most cases, the diffusion-based behavior model achieves slightly better performance. An exception arises in the AntMaze environments under DARE-C, where the policy-based behavior model leads to improved policy performance compared to the diffusion-based instantiation. Tab.~\ref{tab:behavior_model} further summarizes the overall performance of the two behavior model instantiations across task categories.
\begin{table}[H]
\centering
\caption{Comparison of policy-based and diffusion-based behavior model instantiations}
\label{tab:behavior_model}
\renewcommand{\arraystretch}{1.1}
\resizebox{0.44\linewidth}{!}{
\begin{tabular}{l|cc|cc}
\toprule
& \multicolumn{2}{c|}{DARE-C} & \multicolumn{2}{c}{DARE-I} \\
& Policy & Diffusion & Policy & Diffusion \\
\midrule
Locomotion (total) & 865.0 & \textbf{878.1} & 812.1 & \textbf{819.8} \\
AntMaze (total)   & \textbf{539.8} & 537.3 & 417.1 & \textbf{433.2} \\
\midrule
Total      & 1404.8 & \textbf{1415.4} & 1229.2 & \textbf{1253.0} \\
\bottomrule
\end{tabular}}
\end{table}


\begin{figure}[ht]
    \centering
    \includegraphics[width=0.45\linewidth]{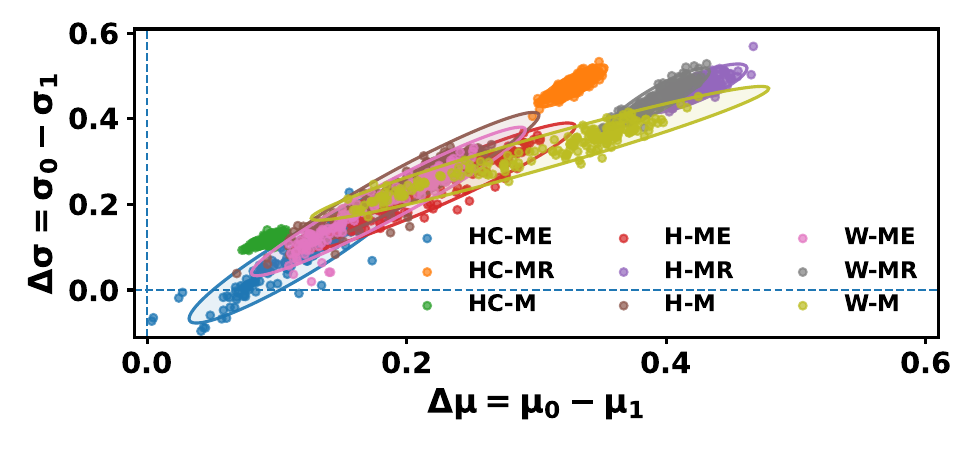}
    \caption{\small Estimated $\mu$ and $\sigma$ gaps during $0.2M$ training steps, both computed using the ongoing policy that represents for online side.}
    \label{fig:statistics_policy}
\end{figure}

We emphasize that DARE does not restrict the behavior model to a specific instantiation and allows alternative choices based on practical considerations. We also show the evolution of the estimated Gaussian statistics in the policy-based case, as shown in Fig.~\ref{fig:statistics_policy}.
\newpage

\section{Additional Results for Reversed Behavior Assignment}
\label{appx:reverse}
\begin{figure}[H]
    \centering
    \includegraphics[width=0.88\linewidth]{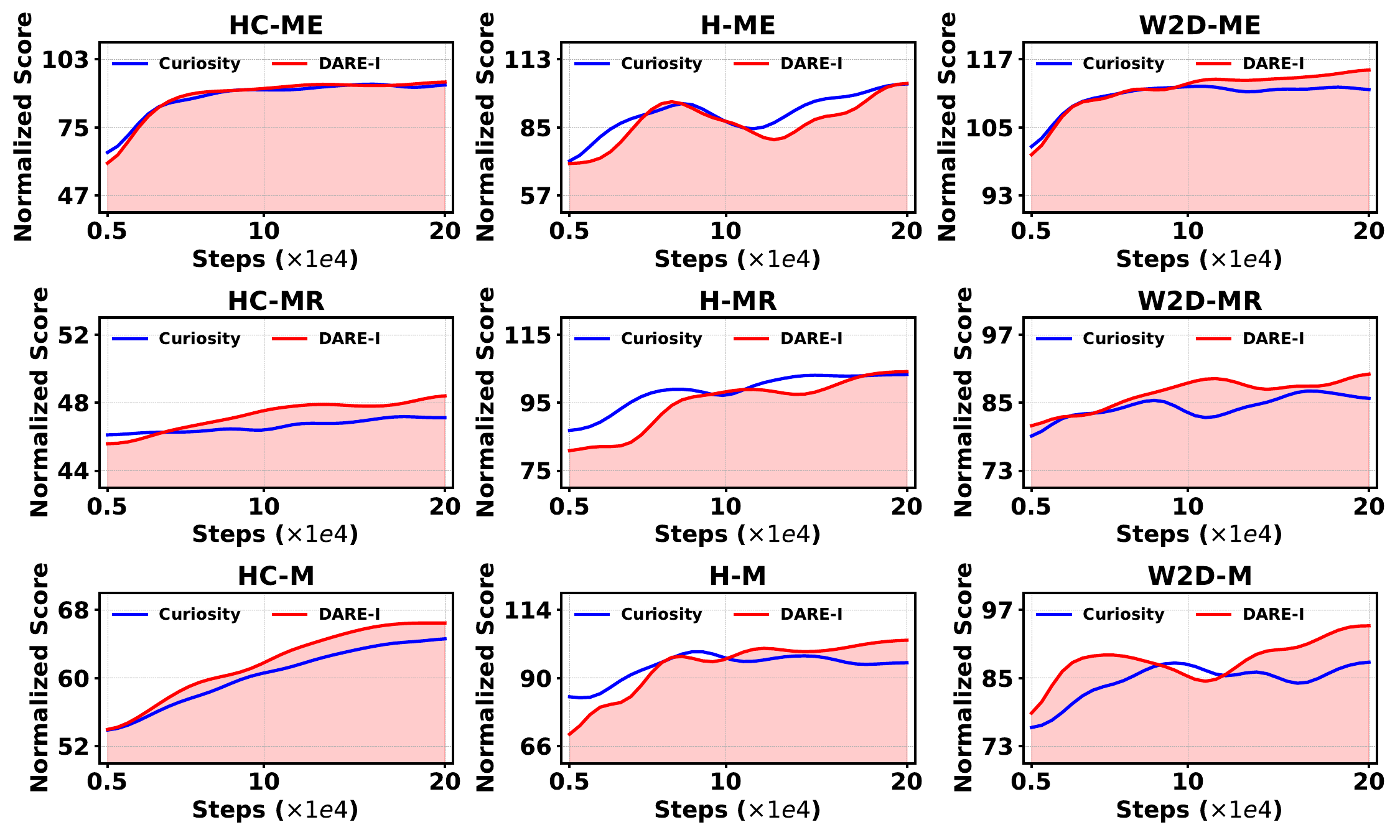}
    \caption{\small Comparison of fine-tuning curves between \texttt{DARE-I.C} (curiosity-style) and \texttt{DARE-I} using a diffusion-based behavior model.}
    \label{fig:c-b}
\end{figure}

\begin{figure}[H]
    \centering
    \includegraphics[width=0.88\linewidth]{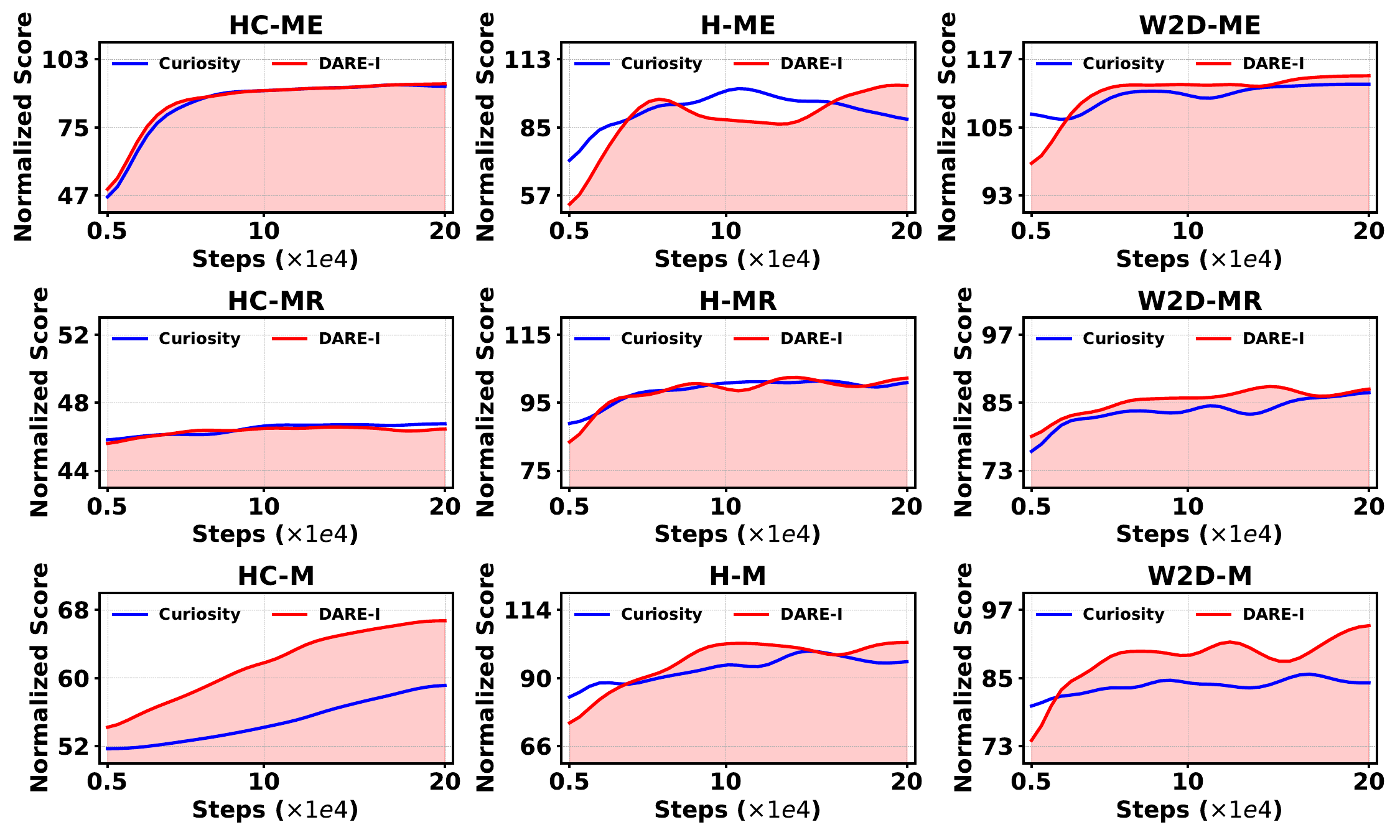}
    \caption{\small Comparison of fine-tuning curves between \texttt{DARE-I.C} (curiosity-style) and \texttt{DARE-I} using a policy-based behavior model.}
    \label{fig:c-p}
\end{figure}

We illustrate the fine-tuning performance curves for \texttt{DARE-I.C} (curiosity-style) and the standard \texttt{DARE-I} under the diffusion-based behavior model in Fig.~\ref{fig:c-b} and the policy-based behavior model in Fig.~\ref{fig:c-p}. Across all MuJoCo locomotion tasks, the reversed assignment (\texttt{DARE-I.C}) is consistently worse than the standard \texttt{DARE-I}, indicating that the inverse exchange operation degrades fine-tuning performance.

\newpage
\section{Additional Ablation Study}
\label{appx:abla}
\begin{figure}[H]
    \centering
    \includegraphics[width=0.88\linewidth]{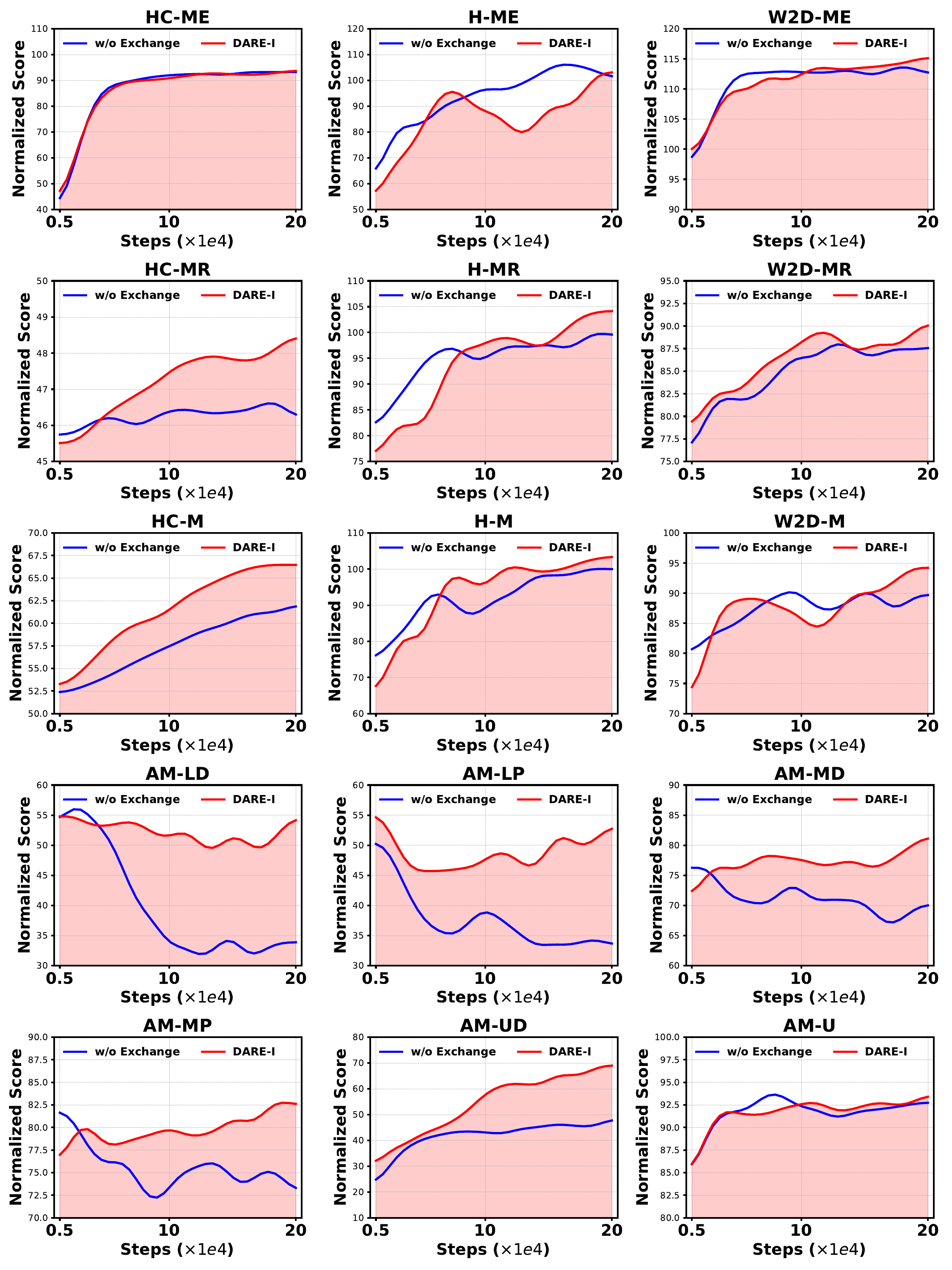}
    \caption{\small Performance curves comparing DARE-I with and without the exchange mechanism.}
    \label{fig:iql_ablation}
\end{figure}

\paragraph{Ablation learning curves.}
In Fig.~\ref{fig:iql_ablation}, we visualize the training trajectories of DARE-I and its w/o Exchange ablation to illustrate the effect of removing the exchange mechanism during training.

\paragraph{Maximum number of exchanges.}
As described in Sec.~\ref{sec:operator}, DARE swaps the same number of samples in the two directions to keep the exchange operation balanced
\[
K = \min\left(
|\mathcal{P}_{\mathrm{off}\rightarrow\mathrm{on}}|,
|\mathcal{P}_{\mathrm{on}\rightarrow\mathrm{off}}|
\right).
\]

In implementation, we optionally impose an upper bound on the number of exchanged samples
\[
K = \min\left(
|\mathcal{P}_{\mathrm{off}\rightarrow\mathrm{on}}|,
|\mathcal{P}_{\mathrm{on}\rightarrow\mathrm{off}}|,
K_{\max}
\right).
\]
This option controls the maximum number of exchanged samples in each mini-batch and prevents unusually large exchanges from dominating a single update. We further examine the effect of this cap on Locomotion tasks. Setting $K_{\max}=32$ achieves a total score of 877.1, while setting $K_{\max}=128$, which is effectively uncapped under our batch setting, achieves 878.2. This observation is also consistent with Fig.~\ref{fig:exchange-apd}, where the number of exchanged samples typically stays around 30.

\paragraph{Behavioral distance choices.}
In DARE, the $L_2$ distance is used as a scalar behavioral deviation statistic between the action $a$ and the behavior reference $a_b(s)$, rather than as a full probabilistic discrepancy measure. For continuous-control O2O RL, $L_2$ provides a simple and parameter-free measure of action deviation, which is sufficient for sample ordering and posterior-based assignment. This choice also keeps the evaluation focused on the contribution of DARE itself, rather than introducing additional complexity through metric engineering. Nevertheless, we further compare DARE-I with alternative distance measures, including $L_1$ distance and Mahalanobis distance. As shown in Tab.~\ref{tab:distance-choices}, the results suggest that DARE is not overly sensitive to the specific behavioral distance measure used in DARE.
\begin{table}[H]
\centering
\caption{Results of DARE-I with different behavioral distance measures.}
\label{tab:distance-choices}
\begin{tabular}{lccc}
\toprule
Locomotion & $L_1$ & Mahalanobis & $L_2$ (Ours) \\
\midrule
M (total)  & 267.3 & 261.9 & 264.5 \\
MR (total) & 238.5 & 237.1 & 242.3 \\
\bottomrule
\end{tabular}
\end{table}

\paragraph{Runtime comparison.}
We further evaluate the runtime on \texttt{hopper-medium-replay-v2}, with all experiments conducted on the same NVIDIA GeForce RTX 5090 GPU. As shown in Tab.~\ref{tab:runtime-cost}, DARE-C (Policy) incurs only a small overhead compared with Cal-QL, while being substantially faster than EDIS-C. DARE-C (Diffusion) requires additional computation due to its diffusion-based behavior modeling component. Since DARE does not require the behavior reference to be parameterized by a diffusion model, the policy-based variant provides a more computationally efficient instantiation when runtime overhead is a concern. 

\begin{table}[H]
\centering
\caption{Runtime comparison on \texttt{hopper-medium-replay-v2}.}
\label{tab:runtime-cost}
\begin{tabular}{lcccc}
\toprule
 & Cal-QL & EDIS-C & DARE-C (Diffusion) & DARE-C (Policy) \\
\midrule
Runtime & 0.5h & 1.6h & 2.1h & 0.6h \\
\bottomrule
\end{tabular}
\end{table}

\end{document}